\documentclass[journal, dvipsnames]{IEEEtran}
\usepackage{cite}
\usepackage{amsmath,amssymb,amsfonts}
\usepackage{algorithmic}
\usepackage{graphicx}
\usepackage{textcomp}
\usepackage{xcolor}
\def\BibTeX{{\rm B\kern-.05em{\sc i\kern-.025em b}\kern-.08em
    T\kern-.1667em\lower.7ex\hbox{E}\kern-.125emX}}

\usepackage{diagbox}
\usepackage{algorithm}
\usepackage{algorithmic}
\usepackage{pgfplots}
\usepackage{multirow}
\usepackage{array}

\usepackage{times}
\usepackage{epsfig}
\usepackage{graphicx}
\usepackage{amsmath}

\usepackage{array}
\usepackage{multirow}
\usepackage[switch]{lineno}
\usepackage{comment}
\usepackage{dsfont}
\usepackage{colortbl}
\usepackage[normalem]{ulem}

\newcommand{\PreserveBackslash}[1]{\let\temp=\\#1\let\\=\temp}
\newcolumntype{C}[1]{>{\PreserveBackslash\centering}p{#1}}
\newcolumntype{R}[1]{>{\PreserveBackslash\raggedleft}p{#1}}
\newcolumntype{L}[1]{>{\PreserveBackslash\raggedright}p{#1}}
\pgfplotsset{
	compat=newest,
	/pgfplots/legend image code/.code={%
		\draw[mark repeat=2,mark phase=2,#1]
		plot coordinates {
			(0cm,0cm)
			(0.18cm,0cm)
			(0.36cm,0cm)
		};
	},
}

\def\eg{\emph{e.g.}} 

\def\ie{\emph{i.e.}}
\def\wrt{\emph{w.r.t.}}
\ifCLASSINFOpdf
\else
\fi

\begin{document}
%
\title{Beyond Universal Person Re-ID Attack}
%

%
\author{
\IEEEauthorblockN{Wenjie~Ding, Xing~Wei,
Rongrong~Ji,~\IEEEmembership{Senior Member,~IEEE},
Xiaopeng~Hong$^\S$,
Qi~Tian,~\IEEEmembership{Fellow,~IEEE},
Yihong~Gong,~\IEEEmembership{Fellow,~IEEE}}
\\
\thanks{Wenjie~Ding is with the College of Artificial Intelligence, Xi’an Jiaotong University, Xi'an 710049, China}
\thanks{Xing~Wei and Yihong~Gong are with the College of Software Engineering, Xi’an Jiaotong University, Xi'an 710049, China}
\thanks{Xiaopeng~Hong is with the School of Cyber Science and Engineering, Xi’an Jiaotong University, Xi'an 710049, China} 
\thanks{Rongrong~Ji is with the Department of Artificial Intelligence, School of Information, Xiamen University, Xiamen, China} 
\thanks{Qi~Tian is with Cloud \& AI, Huawei Technologies, China} 
\thanks{\S \  Corresponding author} 
}

\maketitle

\begin{abstract}
	Deep learning-based person re-identification (Re-ID) has made great progress and achieved high performance recently. In this paper, we make the first attempt to examine the vulnerability of current person Re-ID models against a dangerous attack method, \ie, the universal adversarial perturbation (UAP) attack, which has been shown to fool classification models with a little overhead. We propose a \emph{more universal} adversarial perturbation (MUAP) method for both image-agnostic and model-insensitive person Re-ID attack. Firstly, we adopt a list-wise attack objective function to disrupt the similarity ranking list directly. Secondly, we propose a model-insensitive mechanism for cross-model attack. Extensive experiments show that the proposed attack approach achieves high attack performance and outperforms other state of the arts by large margin in cross-model scenario. The results also demonstrate the vulnerability of current Re-ID models to MUAP and further suggest the need of designing more robust Re-ID models.
\end{abstract}


%
\IEEEpeerreviewmaketitle

\section{Introduction}
%
%
%
%
\IEEEPARstart{I}{n} the past few years, deep learning-based person re-identification (Re-ID) has made remarkable progress and achieved high performance~\cite{8825844, bag-of-tricks,mgn,st-ReID}. Despite the great success, the vulnerability of neural networks has attracted increasing attention in recent years in various applications~\cite{9121297,8792120,MI-FGSM}. It is shown that the output of deep neural networks can be easily attacked by a small perturbation on the input~\cite{FGSM}. After that, many methods have been proposed to attack deep learning-based classification systems~\cite{I-FGSM,MI-FGSM,uap,gduap}. {Classification attack can be accomplished by disturbing decision boundaries which separate the feature space of different classes. However, it brings new, extra challenges to attack an open-set, similarity-ranking problem like person Re-ID. In Re-ID attack, there is no decision boundary in the feature space and the entire similarity rank should be disordered.}
	\begin{figure*}[thbp!]
		\centering
		\small
		\begin{tabular}{c}
			\includegraphics[width=17cm, height=5.19cm]{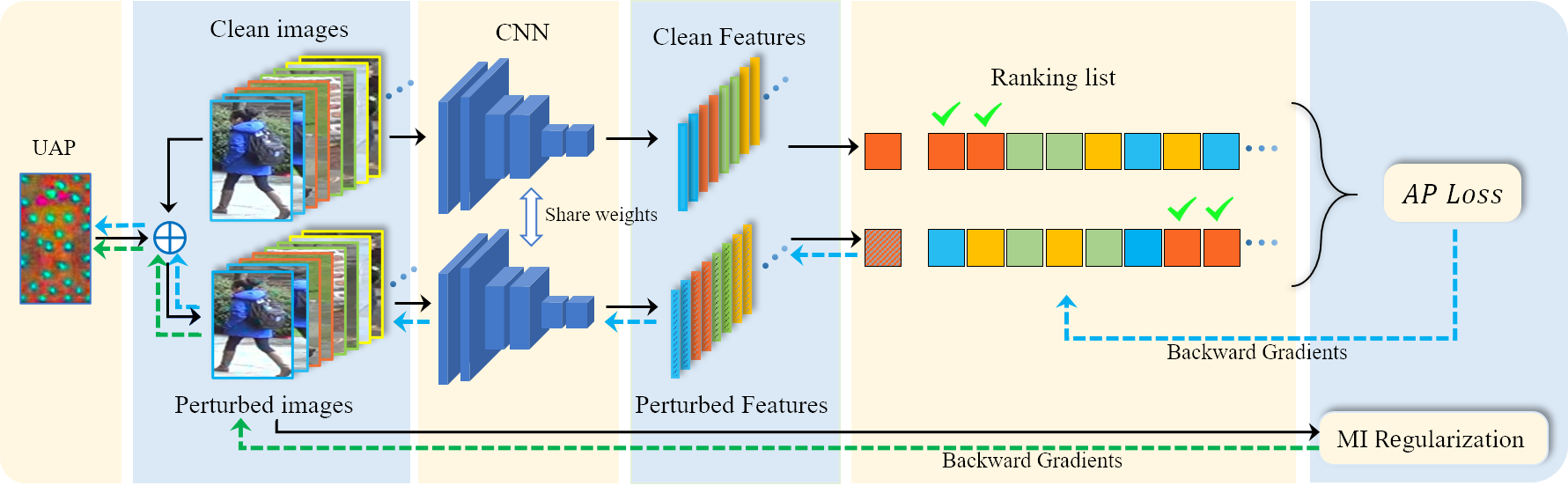} \\
		\end{tabular}
		\caption{\textbf{The learning paradigm of the proposed method.} Different colors in the ranking list and features stand for different IDs and a texture-rendered feature vector indicates an attack. The model weights are fixed and only the perturbation is updated during optimization. The model-insensitive~(MI) regularization is applied on the perturbed image to directly rectify the gradients and constrain the perturbation.} 
		\label{fig:networks}
	\end{figure*}
	
	There are a few works on the adversarial robustness with respect to person Re-ID~\cite{bai2019,zheng2018,ganattack,physical,tripletssim}. However, existing studies on the vulnerability of Re-ID models under adversarial attack faces two problems. Firstly, existing adversarial Re-ID attacks are image-specific~\cite{bai2019,zheng2018,ganattack,physical,tripletssim} and tailored adversarial samples for Re-ID attacks have to be generated for every input image. Secondly, current attack methods usually assume that the adversarial perturbations are generated from and applied to the same model~\cite{bai2019,zheng2018}. As a result, the attack performance dramatically drops when perturbations are applied cross different models. 
	
	{To mitigate the first problem}, Universal Adversarial Perturbation~(UAP)~\cite{uap} is proposed to fool a given deep model using a fixed perturbation map. It is shown that such an image-agnostic approach can achieve high attack performance in classification tasks~\cite{uap,iccv_prior,gduap,fastfeaturefool}. UAP is appealing as it has several advantages over image-specific attack. First of all, UAP is computationally efficient and easy to implement. Once a UAP is trained, the attack can be performed by simply adding the UAP to the input image without expensive online optimization. Second, UAP is more dangerous to the person Re-ID system since it does not require any prior on the appearance of persons.
	{Nonetheless, most of the existing UAP methods focus on the closed-set classification tasks and few of them are specifically designed for the person Re-ID tasks. An effort is thus made in this study to bridge this gap.}
	
	For the second problem, it is clear that in a more realistic scenario, this \emph{model-dependent} problem should be addressed to realize `more' universal, model-insensitive attack. {It is, however, arduous {to disentangle the perturbations from particular model structures, as a certain CNN model is usually required to generate perturbations through the back propagation algorithm. As a consequence, this  problem} remains basically unsolved by the existing UAP attack methods. As we will empirically show later in Section~\ref{subsec:comp_sota}, the cross-model attack performance falls noticeably when the adversarial samples are trained on a source model and used to attack another target model with different CNN structures.
	Although there are several attempts, such as the transferable attack~\cite{iclrtransfer,DBLP:journals/corr/TramerPGBM17,zhou2018transferable} or black-box attack~\cite{narodytska2017simple,Shi_2019_CVPR,Dong_2019_CVPR,Brunner_2019_ICCV} in the literature, they are usually studied in an image-specific attack scenario solely and the overall cross-model attack performance is still far from satisfactory.}

    	In this paper, we make the first attempt to address these two problems jointly for person Re-ID attack. We propose a \emph{more  universal} adversarial perturbation (MUAP) based person Re-ID attack method, which is both image-agnostic and model-insensitive. The MUAP method consists of two key components: 1) a list-wise attack objective to directly attack the global ranking list; and 2) a model-insensitive~(MI) regularization method to consolidate the attack transferability against different model structures and domains. The proposed method is demonstrated in Fig.~\ref{fig:networks}. On the one hand, the list-wise attack objective function aims at disrupting the entire rank of feature similarity and directly decreasing the mean Average Precision (mAP) performance for a Re-ID system. On the other hand, {the model-insensitive regularization rectifies the model gradients to retain a natural image gradient distribution. As a result, {it counteracts the detrimental effect of particular model structures on  model-insensitive universal attack.}}

	{We perform extensive experiments with five different CNN architectures on two widely used Re-ID datasets to evaluate the effects of the proposed method on list-wise attack and cross-model cross-dataset attack. The results clearly show that the proposed MUAP person Re-ID method achieves high attack performance crossing different models and datasets. }
	
	In summary, the contributions of this paper are manifold:
	\begin{itemize}
		\item {We propose a \emph{more universal} adversarial sample generation method for both image-agnostic and model-insensitive person Re-ID attack.}
		\item {We devise an AP loss based list-wise attack objective function to efficiently disrupt the entire rank list according to the similarity between training samples.}
		\item  {We propose a {simple yet super-efficient} model-insensitive regularization term for
		universal attack against different CNN structures.}
	\end{itemize}
	
	\renewcommand{\tabcolsep}{1.0pt}

\section{Related Work}

	{\flushleft\textbf{Adversarial attack.}} Lots of methods have been proposed to effectively attack the neural network based models~\cite{I-FGSM,MI-FGSM,score_based1,decision_based1}. Adversarial attacks pose threat not only to the computer vision tasks~\cite{8621025,score_based1}, but also to other tasks adopting machine learning models, such as malware detection~\cite{9121297}, wireless communications~\cite{8792120} and audio-related tasks~\cite{8949445, 8922608}. One-step gradient-based method~\cite{FGSM} is proposed to generate adversarial samples by maximizing the network’s prediction error for one single step. I-FGSM~\cite{I-FGSM} further propose to update gradients iteratively. MI-FGSM ~\cite{MI-FGSM} considers the direction of previous optimization step and claims better performance. These gradient-based methods generate perturbations based on back-propagation gradients, but there are cases where the target model parameters could not be accessed. Since adversarial perturbations trained on one model can usually fool other models~\cite{FGSM}, a typical way in this scenario is to train perturbation on a source model and use it to attack the target model, which is usually referred to as transfer attack or black-box attack~\cite{iclrtransfer,zhou2018transferable}. Besides the gradient-based methods, there are also query-based ones which train the perturbation iteratively based on the model outputs of the query images~\cite{Andriushchenko2019SquareAA,9152788}. Most of the attack methods are image-specific. More recently, Moosavi t al.~\cite{uap} showed the existence of universal image-agnostic perturbations. Mopuri \textit{et al.}~\cite{gduap} propose a task-independent objective to train more generalizable UAP. 
	
	{The open-set tasks like person Re-ID are {different} from the classification tasks in nature and thus it is infeasible to straightforwardly apply existing attack methods against such classification tasks to person Re-ID. Firstly, in classification tasks, a decision boundary to distinguish different classes can be formed within the feature space once a certain type of classifiers is chosen. A classification attack is considered as a success if the input polluted by the noise results in a wrong label. In contrast, in Re-ID, there is no class boundary in the feature space. As a result, a successful attack should disorder the whole global ranking list, not just ruin the Top-1 results. Secondly, in classification, once a model is learned, its output is only determined by the input image. However, the output of Re-ID, \ie, the ranking list returned by Re-ID system, is determined by not only the input query image but also the gallery set~\cite{lijie}. Therefore, the Re-ID attack brings new challenges.}
	
	{\flushleft\textbf{Adversarial attack against person Re-ID.}} {There are a couple of recent studies on adversarial attack against person Re-ID.
		Bai~\textit{et al.}~\cite{bai2019} propose an Adversarial Metric Attack (AMA) method to generate adversarial gallery samples by maximizing the distance between clean images and perturbed images. Opposite-Direction Feature Attack (ODFA) is designed to generate perturbations by pushing the perturbed images away in the opposite direction of raw images in feature space~\cite{zheng2018}. In~\cite{ganattack}, query-specific perturbations are produced through an unsupervised learning approach. Wang~\textit{et al.}~\cite{physical} generate a physical-world adversarial pattern on clothes for each person. Wang \textit{et al.}~\cite{tripletssim} propose to generate adversarial 
	queries using GAN-based network. Li \textit{et al.}~\cite{lijie} propose retrieval-against UAP (IR-UAP) which can be adapted to Re-ID attack.}
	
	{This study is distinct from these existing ones about Re-ID attack. Firstly, we focus on a difficult problem of producing an \emph{image-agnostic, universal} noise, while the existing methods usually work on image-specific noises~\cite{bai2019, ganattack, zheng2018, physical}. Secondly and more importantly, we fulfill the more advanced requirement of Re-ID UAP attack, \ie, the \emph{model-insensitive} requirement, while such a cumbersome requirement is usually out of the scopes of the existing UAP attack~\cite{lijie, uap, iclrtransfer,zhou2018transferable, Shi_2019_CVPR} and Re-ID attack studies~\cite{bai2019, ganattack, zheng2018, physical}. The proposed regularization term is related to total variation~\cite{rudin1992nonlinear} in form. Total variation is previously used for image denoising, but here we introduce it into the attack process to produce more transferable UAPs.}

	\section{Methodology}
	In this section, we present the proposed methods to generate the \emph{more universal} adversarial perturbation against Re-ID models. We first give problem definition, and then introduce the list-wise attack objective and the proposed model-insensitive regularization method. 
	\subsection{Framework}
	
	{\flushleft\textbf{Problem definition.}} Let $\mathcal{D} \subset \mathbb{R}^m$ be a database of pedestrian images, and $f_{\theta}$ be a given Re-ID model, where $\theta$ denotes model parameters, \eg, a CNN which embeds an image $x \in \mathcal{D}$ to a feature vector space $f_{\theta}(x) \in \mathbb{R}^n$. Typically, the model has been trained with the objective that the distances of features with the same person IDs (positive pairs) are smaller than those with different IDs (negative pairs). The goal of UAP attack is to seek a perturbation vector \textbf{\textit{u}} with small magnitudes that fools the model $f_{\theta}$ so that for almost all query samples in $\mathcal{D}$, the negatives rank above the positives. We denote the evaluation metric of Re-ID results as $M$. UAP attack against Re-ID aims to find a vector $u \in \mathbb{R}^m$ which satisfies the following constraints:
	\begin{equation}
	    u = \mathop{\arg\min}_{u} M , \; \textit{s.t.} {\left\| u \right\|_\gamma} \le \epsilon
	\end{equation}
	where ${\left\| {\cdot} \right\|_\gamma}$ represents the $\ell_\gamma$ norm, and $\epsilon $ limits the magnitude of the perturbation vector $u$. As the gallery database is usually quite large and not available in attack, we consider a more practical situation where only the query image is attacked. 
	{\flushleft\textbf{Baseline.}} We build the baseline for Re-ID attack by adopting the iterative least likely class method~\cite{I-FGSM}, which is widely used for attack against classification. The iterative least likely class method tries to make the perturbed image $q$ classified to the least likely class $y_{LL}=\arg \min_{y} \{p(y|q)\}$. The objective function to minimize is formulated as follows:
	\begin{equation}
	\mathcal{L}_{Base} = - \log (p(y_{LL}|q+u)).
	\end{equation}
	
	The classification-against attack methods consider single images and ignore their relation. However, the relation between images, especially the relation between the query images and the gallery ones is fundamental and determines the Re-ID results, \ie the ranking list returned by Re-ID systems. Therefore, instead of adopting the classification-against methods to Re-ID attack, we propose to attack the Re-ID system by disrupting the entire similarity rank.

	{\flushleft\textbf{Overall Objective.}}
	The objective function of our method is composed of a base list-wise attack objective for general image-agnostic attack and a model-insensitive regularizer to enhance cross-model attack performance. The list-wise objective aims at disrupting the entire rank of feature similarity. And the model-insensitive regularizer aims to regularize the perturbation during training to avoid overfitting and biasing on the specific model. The overall objective is formulated as follows:
	\begin{equation}
	\mathcal{L} = \mathcal{L}_{AP} + \lambda \cdot \mathcal{L}_{MI},
	\label{eq:all}
	\end{equation}
	where $\mathcal{L}_{AP}$ and $\mathcal{L}_{MI}$ represent the attack objective and the model-insensitive regularization respectively, and $\lambda$ is a parameter balancing this two terms.

	\subsection{{List-wise Attack Objective}}
	For a query person image $q\in \mathcal{D}$, let $q'=q+u$ be the attacked image, $\mathcal{D}_q^+$ and $\mathcal{D}_q^-$ be the positive and negative gallery set with respect to $q$. Given a distance metric $d(\cdot,\cdot)$, let $\mathcal{L}_{q'} = (x_1,x_2,\cdots,x_l)$ be the ranking list \wrt \,  $q'$ in $\mathcal{D}_q^+ \cup \mathcal{D}_q^-$ in ascent order, \ie, $d(f_{\theta}(q'),f_{\theta}(x_1)) \le d(f_{\theta}(q'),f_{\theta}(x_2)) \le \cdots \le d(f_{\theta}(q'),f_{\theta}(x_l))$, where $l = |\mathcal{D}_q^+ \cup \mathcal{D}_q^-|$ is the length of the list.
	The average precision (AP) of the ranking is defined by the average of precision values evaluated at each rank position:
	\begin{equation}
	AP = {\frac{1}{|\mathcal{D}_q^+|} \sum_{k=1}^l{P_k \cdot \mathds{1}[x_k \in \mathcal{D}_q^+]}},
	\label{eq:ap}
	\end{equation}
	where $\mathds{1}[\cdot]$ is the binary indicator function, $|\cdot|$ denotes cardinality, and $P_k$ denotes precision at the $k$-th position which is formulated as:
	\begin{equation}
	P_k = \frac{1}{k} \sum_{k'=1}^k{\mathds{1}[x_{k'} \in \mathcal{D}_q^+]},
	\end{equation}
	$AP \in [0,1]$ and the maximum value is achieved if and only if every instance in $\mathcal{D}_q^+$ ranks above all instances in $\mathcal{D}_q^-$. 
	Finally, mean average precision (mAP) accumulates AP in a dataset and gives their average
	\begin{equation}
	mAP = \frac{1}{|\mathcal{D}|}\sum_{q \in \mathcal{D}}{AP}.
	\label{eq:map}
	\end{equation}

	Due to the sorting operation, AP is non-smooth and non-differential. Thus it is difficult to optimize AP in neural networks using the standard back-propagation training method. Here we adopt the soft histogram approximation~\cite{retrieval-rank}, where the non-differential sorting operation is replaced by histogram binning and soft indication.
	We assume the CNN embedding $f_{\theta}(\cdot)$ is $\ell_2$ normalized, and let $d(f_{\theta}(q'),f_{\theta}(x_i))$ be the Euclidean distance of CNN embeddings of the attacked query and gallery image which lies in $[0,2]$. Given a histogram bin number $b$ and equally divide the distance interval into $b$-$1$ parts with length $\Delta=\frac{2}{b-1}$. Similar to~\cite{retrieval-rank,hashing-rank}, a soft indicator function $\delta:\mathbb{R}\times\{1,2,\cdots,b\}\rightarrow[0,1]$ is defined, thus the contribution of the each gallery instance $x_i$ to the $k$-th bin ($k\in[1,b]$) is calculated by:
	
	\begin{equation}
	\delta(x_i, k) = max\left(1-\frac{\| d(f_{\theta}(q'),f_{\theta}(x_i)) - (k-1)\Delta \|_1}{\Delta}, 0\right),
	\label{eq:delta}
	\end{equation}
	where $\|\cdot\|_1$ is the $\ell_1$ norm. In this way, we can calculate precision at each bin instead of at each rank position to avoid the sorting operation. The precision at the $k$-th bin is:
	\begin{equation}
	\hat{P}_k = \frac{\sum_{k'=1}^k\sum_{i=1}^l{\delta(x_i,k') \cdot \mathds{1}[x_i\in \mathcal{D}_q^+]}}{\sum_{k'=1}^k\sum_{i=1}^l{\delta(x_i,k')}}.
	\label{eq:app_prec}
	\end{equation}

	In an analogy to Eqs. 4 and 5, we can define the approximated average precision $\hat{AP}$.
	Finally, {we define our list-wise loss function to minimize on top of $\hat{AP}$ as follows.}
	\begin{equation}
	 \mathcal{L}_{AP} = \frac{1}{|\mathcal{D}_q^+|} \sum_{k=1}^b \hat{P}_k \cdot \left(\sum_{i=1}^l {\delta(x_i,k)\cdot \mathds{1}[x_i \in \mathcal{D}_q^+]}\right).
	\label{eq:app_ap}
	\end{equation}
	
\setlength{\fboxrule}{0.3pt}
	\setlength{\fboxsep}{0in}
	\begin{figure*}[!t]
		\small
		\centering
		\renewcommand{\tabcolsep}{1.0pt}
		\renewcommand\arraystretch{0.5}
		\begin{tabular}{ccccccc}
			\fbox{\includegraphics[height=0.18\textwidth]{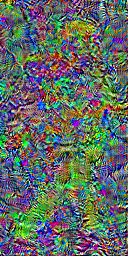}} &
			\fbox{\includegraphics[height=0.18\textwidth]{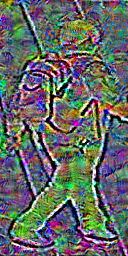}} &
			\begin{tikzpicture}
			\begin{axis}[width=0.09\textwidth,height=0.18\textwidth,scale only axis,ymin=0,ymax=256,xmin=0,xmax=128,ticks=none,
			]
			\addplot[thick,blue] graphics[xmin=0,ymin=0,xmax=128,ymax=256] {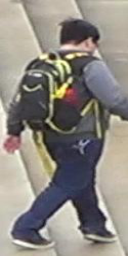};
			\draw [thick, draw=Goldenrod] (axis cs: 0,50) -- (axis cs: 128,50);
			\end{axis}
			\end{tikzpicture} &
			\begin{tikzpicture}
			\begin{axis}[width=0.09\textwidth,height=0.18\textwidth,scale only axis,ymin=0,ymax=256,xmin=0,xmax=128,ticks=none,
			]
			\addplot[thick,blue] graphics[xmin=0,ymin=0,xmax=128,ymax=256] {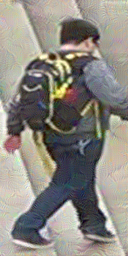};
			\end{axis}
			\end{tikzpicture} &
			\begin{tikzpicture}
			\begin{axis}[width=0.09\textwidth,height=0.18\textwidth,scale only axis,ymin=0,ymax=256,xmin=0,xmax=128,ticks=none,
			]
			\addplot[thick,blue] graphics[xmin=0,ymin=0,xmax=128,ymax=256] {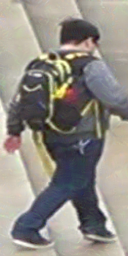};
			\end{axis}
			\end{tikzpicture} &
			\includegraphics[height=0.1834\textwidth]{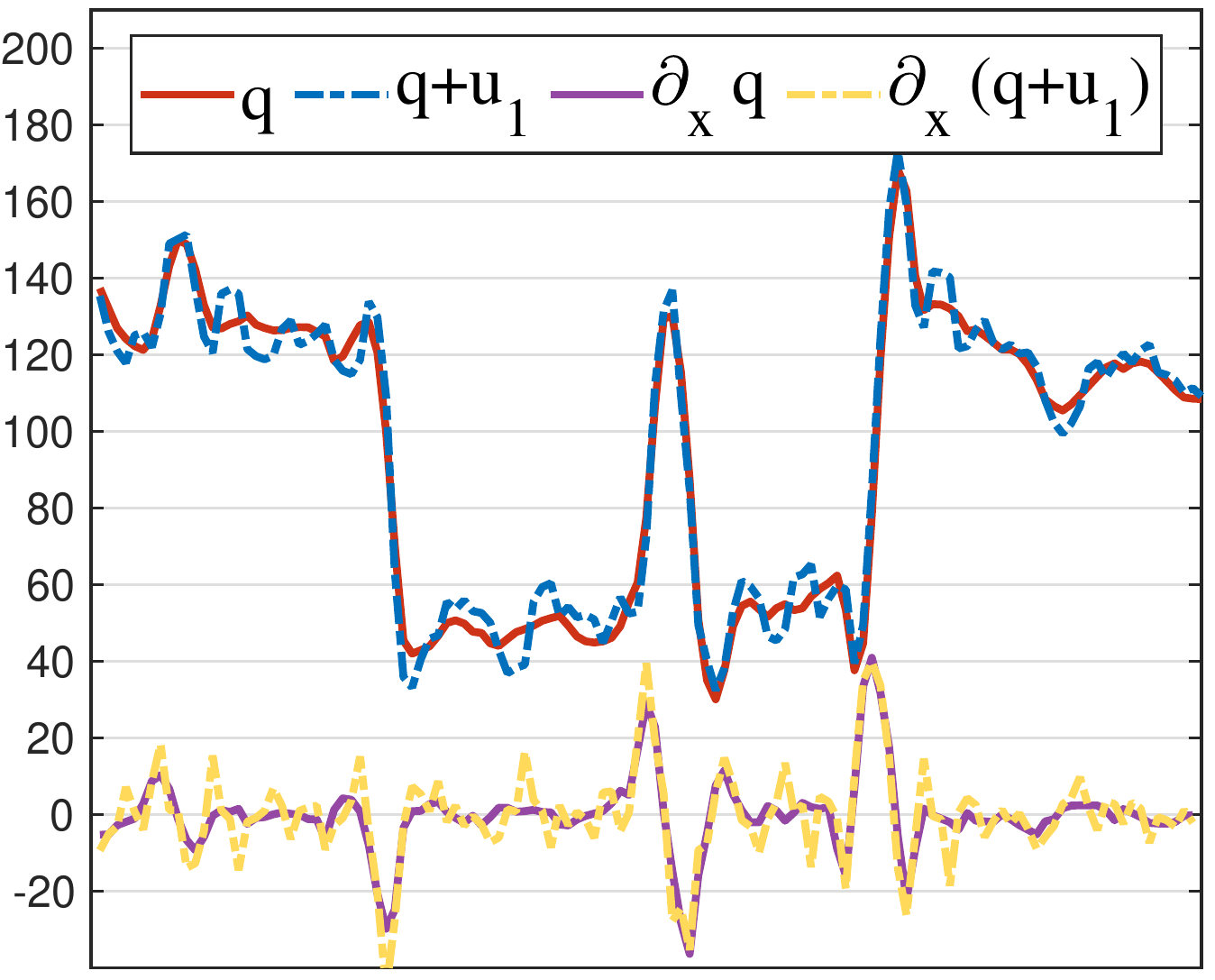} &
			\includegraphics[height=0.1834\textwidth]{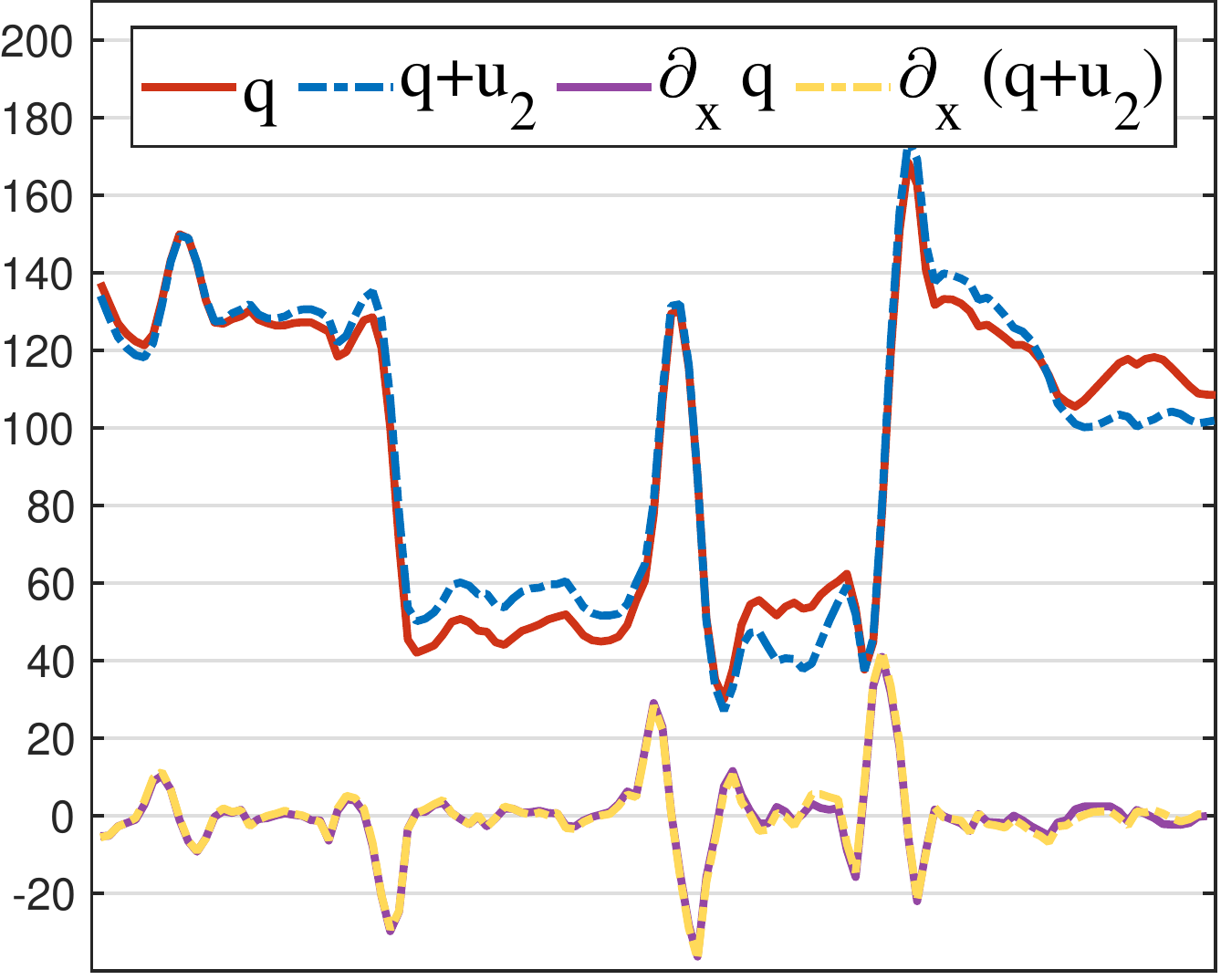} \\
			\fbox{\includegraphics[width=0.09\textwidth]{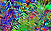}} &
			\fbox{\includegraphics[width=0.09\textwidth]{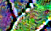}} &
			\fbox{\includegraphics[width=0.09\textwidth]{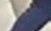}} &
			\fbox{\includegraphics[width=0.09\textwidth]{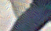}} &
			\fbox{\includegraphics[width=0.09\textwidth]{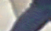}} &
			\,\,\,\,\,\fbox{\includegraphics[width=0.21\textwidth]{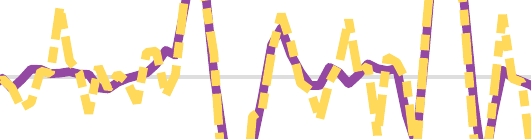}} &
			\,\,\,\,\,\fbox{\includegraphics[width=0.21\textwidth]{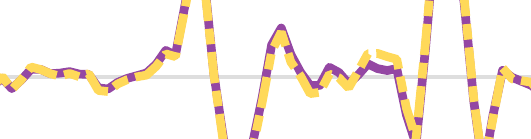}} \\
			\vspace{2mm}
			(a) $u_1$ & (b) $u_2$ & (c) $q$ & (d) $q+u_1$ & (e) $q+u_2$ & (f) & (g) \\
		\end{tabular}
		\caption{\textbf{Comparison of perturbations trained with and without the model-insensitive regularization.} (a)-(b): Two adversarial perturbations for an input image during the training procedure, without ($u_1$) and with ($u_2$) regularization. Pixel values are amplified ($10\times$) for visualization. (c)-(e): The input and two attacked images. (f)-(g): Pixel values $q$ for a channel of a scan line indicated in (c), gradient values $\partial_x q$, attacked pixel values $q+u_1$ and $q+u_2$, and their gradients $\partial_x (q+u_1)$ and $\partial_x (q+u_2)$. }
		\label{fig:example}
	\end{figure*}
	The approximation method defined in Eqs.~\eqref{eq:app_prec} and \eqref{eq:app_ap} do not rely on the sorting operation, thus $\mathcal{L}_{AP}$ can be optimized in neural networks in an end-to-end manner to directly attack the accuracy of person Re-ID. 
	\begin{figure}
		\centering
		\begin{tikzpicture}
		\begin{axis}[width=0.39\textwidth,height=0.26\textwidth,
		xmin=0.7,xmax=2.3,ymin=500,ymax=1400,
		ybar,
		xtick={1,2},
		xticklabels={{DukeMTMC-reID},{Market-1501}},
		ylabel=Gradient energy,
		enlargelimits=0.15,
		legend style={at={(1.17,1.0)},draw=none,anchor=north},
		legend cell align=left,
		legend columns=1,
		bar width=16pt,
		nodes near coords,
		nodes near coords align={vertical},
		every node near coord/.append style={font=\tiny},
		font=\small,
		]
		\addplot[color=BrickRed,fill] coordinates {(1,941.9) (2,677.2)};
		\addplot[color=NavyBlue,fill]  coordinates {(1,1292.7) (2,1146.4)};
		\addplot[color=Dandelion,fill]  coordinates {(1,954.2) (2,695.7)};
		\legend{$q$,$q+u_1$,$q+u_2$}
		\end{axis}
		\end{tikzpicture}
		\caption{Average gradient energy on two datasets, where $q$ represents the original image and $u_1$ and $u_2$ are the UAPs trained without and with the MI regularization, respectively.}
		\label{fig:dataset_tv}  
	    \vspace{-0.5cm}
	\end{figure}
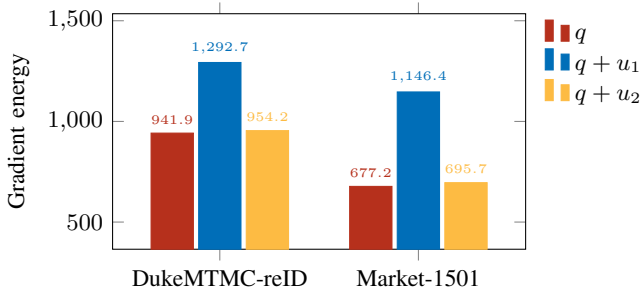 
	\subsection{{Model-Insensitive Regularization}}
	
	Transferability refers to the ability of a UAP trained on a source model to attack a different target model. The transferability of UAPs across different models and different environments is important for person Re-ID attack in practice. However, the transfer attack performance is usually much lower than the white-box attack, which suggests the over-fitting and bias of the UAP to the specific training model.
	{Since both $\mathcal{L}_{base}$ and $\mathcal{L}_{AP}$ depend on the CNN architecture $f$ and its parameter set $\theta$, they are inevitably related to the training model. To achieve the model insensitivity, we define a regularization term that is independent of $f$ and $\theta$ as follows: }
	
	\begin{equation}
	\mathcal{L}_{MI} = \sum_{i=1}^m{\left\|(\nabla q')_i\right\|_p^p} = \sum_{i=1}^m{\left(\left\|(\partial_x q')_i\right\|_p^p +  \left\|(\partial_y q')_i\right\|_p^p\right)},
	\label{eq:tv}
	\end{equation}

\noindent	where $q' = q+u$ denotes the attacked image, $\nabla$ denotes the gradient operation, and ${\left\| {\cdot} \right\|_p}$ denotes the $\ell_p$ norm. The right side of Eq.~\ref{eq:tv} sums the gradients of the perturbed image with respect to two directions and thus reflects the image gradient energy. $\ell_p$ can be any forms of norm, we mainly consider the $\ell_1$ and $\ell_2$ form in this paper. Detail results are reported in \ref{l1l2norm}.
	
	 The introduction of model-insensitive~(MI) regularization brings the following benefits. Firstly, it is clear that Eq.~\ref{eq:tv} is model-insensitive as no information related to model structures and parameter settings is used to define $\mathcal{L}_{MI}$. Secondly, minimizing $\mathcal{L}_{MI}$ punishes the UAP {with unusual pixel changes and helps to avoid artifacts in the high-frequency part of the attacked image.
	{It helps to maintain the gradient distribution of perturbed images as that of natural images, which not only forces the perturbed image to appear \emph{natural}, but also reduces the risk of being interfered with other factors such as particular model structures. }
	Thirdly, minimizing $\mathcal{L}_{MI}$ has the effect on embedding the adversarial perturbations into pixels with high image gradient magnitude. 

	{Psychological studies have suggested that slight variations of the pixels (\eg, caused by an adversarial signal with small energy)  become subtle if they are placed around contours. This phenomenon is referred to as by the \emph{visual masking} effect~\cite{netravali1977adaptive}.  As a result, it significantly increases the difficulty to defend against such unnoticeable adversarial signals.}}

	We compare the perturbations trained with and without $\mathcal{L}_{MI}$ on single image in Fig.~\ref{fig:example}.\,(a)-(b) display two adversarial perturbations for an input image ($q$) during training, without ($u_1$) and with ($u_2$) $\mathcal{L}_{MI}$, (c)-(e) display the input and attacked ($q+u_1$ and $q+u_2$) images, and (f)-(g) display pixel values ($q$) of a scan line, their gradients ($\partial_x q$), attacked pixel values ($q+u_1$ and $q+u_2$), and their gradients ($\partial_x (q+u_1)$ and $\partial_x (q+u_2)$), respectively. As can be seen from (f)-(g), compared to $\partial_x (q+u_1)$, $\partial_x (q+u_2)$ is closer to $\partial_x q$ so that the attacked image is less noisy and more natural when regularized by $\mathcal{L}_{MI}$. The final trained UAPs can be found in Fig.~\ref{fig:moreuap_vis}. 
	As defined in Eq.~\ref{eq:tv}, $\mathcal{L}_{MI}$ is applied on the input image, and independent of both the CNN architecture $f$ and the distribution of its parameter $\theta$. Consequently, $\mathcal{L}_{MI}$ rectifies $q'$ to reduce the risk of biasing $u$ on the model $f_{\theta}$ exclusively. We show later that such a model-insensitive regularization can significantly improve the performance of cross-model attack in Section~\ref{sec:results}. On the other hand, $\mathcal{L}_{MI}$ may also be applied on the perturbation $u$ itself, and thus this term becomes data-free. Nevertheless, in this way inherent structures in data is neglected. Therefore, we enforce the $\mathcal{L}_{MI}$ term to be coupled with the input to explore characteristics of data for improving the attack performance in this work.
	Experimental results in Section~\ref{sec:results} also show that our method generalizes well to different datasets.

	\begin{algorithm}[tb]
		\caption{Learning More Universal Adversarial Perturbations Against Person Re-Identification}
		\label{alg}
		\begin{algorithmic}[1] 
			\REQUIRE Database $\mathcal{D}$, source model $f_{\theta}$,  number of epochs $T$, $\ell_\gamma$ norm of perturbation bound $\epsilon$, $\lambda$ for balancing the $\mathcal{L}_{MI}$ term.
			\ENSURE The universal perturbation vector $u$.
			\STATE Initialize $u \leftarrow 0$,
			\WHILE{epochs $\le T$}
			\FOR{each sample $q \in \mathcal{D}$}
			\STATE Update perturbation vector $u$ using Eq.~\eqref{eq:update}. 
			\IF{$u$ get saturated}
			\STATE Constrain perturbation vector $u$ using the projection defined in Eq.~\eqref{eq:projector}.
			\ENDIF
			\ENDFOR
			\ENDWHILE
		\end{algorithmic}
	\end{algorithm}
	
	$\mathcal{L}_{MI}$ is {the gradient energy which statistically measures of the complexity of an image with respect to its spatial variation.} We calculate average gradient energy of the original and attacked images using $u_1$ and $u_2$ on two datasets, and present the results in Fig.~\ref{fig:dataset_tv}. As it can be seen, $q+u_1$ can increase the gradient energy remarkably, which makes them statistically different from uncontaminated images, while $q+u_2$ maintains the same level as $q$ with an additional perturbation noise. This phenomenon indicates that the preservation of some statistical character during attack may enhance the attack transferability {across models}.

\subsection{{Optimization}}
	Algorithm~\ref{alg} summarizes the overall learning process. We learn the perturbation vector iteratively in mini-batch. At the $t$-th iteration, we compute $\mathcal{L}_{AP}$ by treating each image in the batch as the query image and others as gallery images. Then the gradient of loss function with respect to the perturbation ${\nabla}_u \mathcal{L}$ is calculated using back-propagation. After that, we use the stochastic gradient decent with momentum~\cite{MI-FGSM} to update the perturbation as follows:
	\begin{equation}
	\left\{
	\begin{aligned} 
 	g_t &\leftarrow \eta \cdot g_{t-1} + \frac{{\nabla}_u \mathcal{L}}{\left\| {\nabla}_u \mathcal{L} \right\|_1} \\
	u_{t}   &\leftarrow u_{t-1} + \alpha \cdot sign(g_t) , 
	\end{aligned}\right.,
	\label{eq:update}
	\end{equation}
	where $g_t$ accumulates the gradients of the first $t$ iterations with a decay factor $\eta$, and $\alpha$ is the learning rate. 
	Similar to~\cite{uap}, we define a projection operator to decrease the perturbation if it gets saturated:
	\begin{equation}
    u \leftarrow \arg \min_{u{'}}{\|u-u{'}\|_p}, \,\, s.t.  \|u{'}\|_p \le \epsilon ,
    \label{eq:projector}
    \end{equation}
    where $\epsilon $ limits the magnitude of the perturbation vector $u$.
	\section{Experiments}
	
	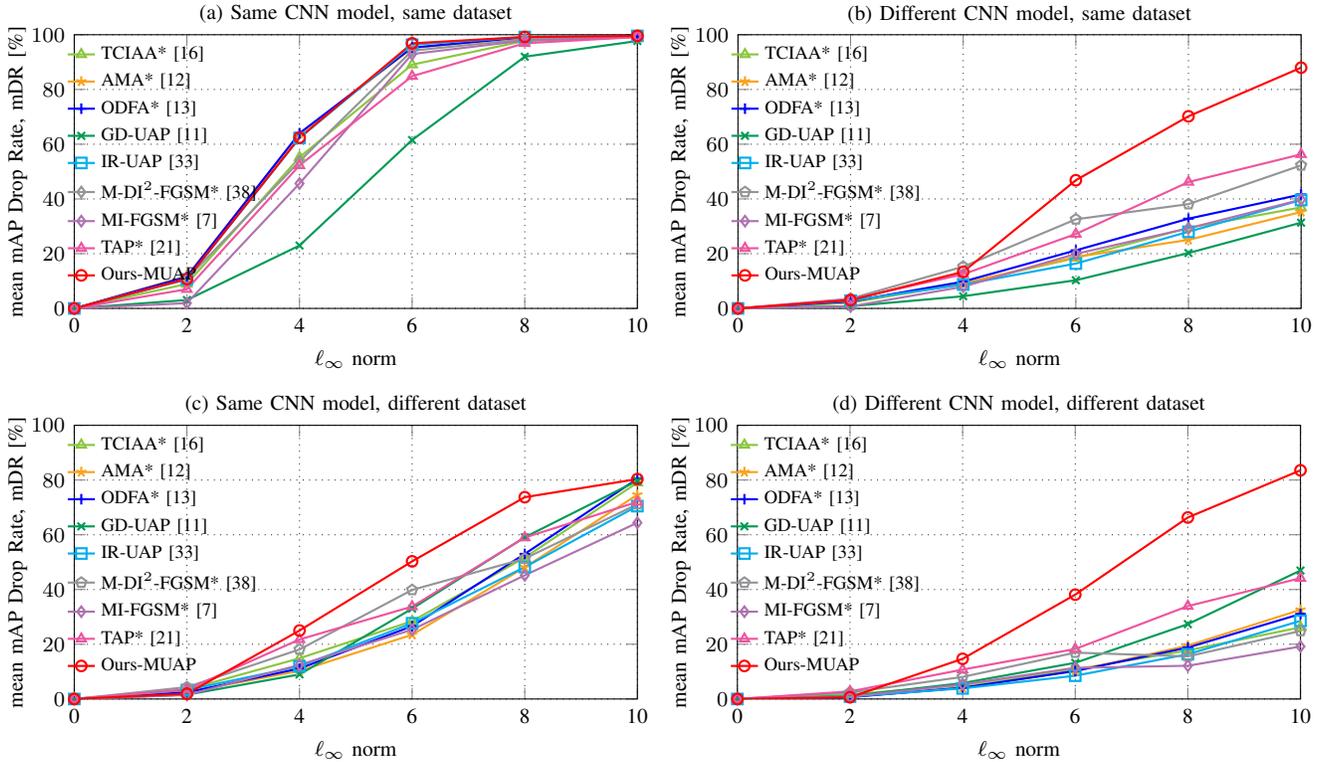
\begin{figure*}[!ht]
		\centering
		\begin{tabular}{cc}
			\begin{tikzpicture}[/pgfplots/width=0.5\linewidth,/pgfplots/height=0.288\linewidth]
			\begin{axis}
			[ymin=0,ymax=100,xmin=0,xmax=10,font=\footnotesize,
			xtick={0,2,4,6,8,10},
			xticklabels={0,2,4,6,8,10},
			title={(a) Same CNN model, same dataset},
			title style={yshift=-1.1ex},
			xlabel={{$\ell_\infty$ norm}},
			ylabel={{mean mAP Drop Rate, mDR [\%]}},
			ylabel style={yshift=-1.2ex},
			legend style={at={(0.16,0.999)},draw=none,anchor=north,legend columns=1,font=\scriptsize,fill=white,fill opacity=0.0,text opacity=1},
			legend cell align={left},
			legend entries={ TCIAA*~\cite{tripletssim},AMA*\cite{bai2019},ODFA*\cite{zheng2018},GD-UAP\cite{gduap},IR-UAP\cite{lijie}, M-DI$^2$-FGSM*~\cite{M-DI-FGSM},MI-FGSM*~\cite{MI-FGSM},  TAP*~\cite{zhou2018transferable},  Ours-MUAP},
			grid=both,
			grid style=dotted,
			major grid style={white!20!black},
			minor grid style={white!70!black}]
			\addplot[LimeGreen,thick,mark=triangle] coordinates{(0,0) (2,8.94) (4,55.22) (6,89.07) (8,97.73) (10,99.13)}; 
			\addplot[YellowOrange,thick,mark=star] coordinates{(0,0) (2,11.84) (4,63.79) (6,95.11) (8,99.01) (10,99.50)};  
			\addplot[blue,thick,mark=+] coordinates{(0,0) (2,11.53) (4,64.01) (6,95.30) (8,99.05) (10,99.47)};  
			\addplot[ForestGreen,thick,mark=x] coordinates{(0,0) (2,3.07) (4,22.981) (6,61.54) (8,91.97) (10,97.71)};  
			\addplot[cyan,thick,mark=square] coordinates{(0,0) (2,10.60) (4,62.24) (6,96.36) (8,99.17) (10,99.56)};  
			\addplot[Gray,thick,mark=diamond] coordinates{(0,0) (2,10.72) (4,53.76) (6,94.18) (8,98.35) (10,99.41)}; 
			\addplot[Orchid,thick,mark=diamond] coordinates{(0,0) (2,1.93) (4,45.65) (6,92.94) (8,97.89) (10,99.26)}; 
			
			\addplot[Rhodamine,thick,mark=triangle] coordinates{(0,0) (2,7.02) (4,52.24) (6,84.86) (8,96.88) (10,99.16)}; 
			
			\addplot[red,thick,mark=o] coordinates{(0,0) (2,10.75) (4,62.42) (6,96.86) (8,99.19) (10,99.59)};  
			\end{axis}
			\end{tikzpicture}&
			\begin{tikzpicture}[/pgfplots/width=0.5\linewidth,/pgfplots/height=0.288\linewidth]
			\begin{axis}
			[ymin=0,ymax=100,xmin=0,xmax=10,font=\footnotesize,
			xtick={0,2,4,6,8,10},
			xticklabels={0,2,4,6,8,10},
			title={(b) Different CNN model, same dataset},
			title style={yshift=-1.2ex},
			xlabel={{$\ell_\infty$ norm}},
			ylabel={{mean mAP Drop Rate, mDR [\%]}},
			ylabel style={yshift=-1.1ex},
			legend style={at={(0.16,0.999)},draw=none,anchor=north,legend columns=1,font=\scriptsize,fill=white,fill opacity=0.0,text opacity=1},
			legend cell align={left},
			legend entries={TCIAA*~\cite{tripletssim},AMA*\cite{bai2019},ODFA*\cite{zheng2018},GD-UAP\cite{gduap},IR-UAP\cite{lijie}, M-DI$^2$-FGSM*~\cite{M-DI-FGSM},MI-FGSM*~\cite{MI-FGSM},  TAP*~\cite{zhou2018transferable},  Ours-MUAP},
			grid=both,
			grid style=dotted,
			major grid style={white!20!black},
			minor grid style={white!70!black}]
 			\addplot[LimeGreen,thick,mark=triangle] coordinates{(0,0) (2,2.07) (4,9.05) (6,18.44) (8,29.40) (10,36.91)}; 
			\addplot[YellowOrange,thick,mark=star] coordinates{(0,0) (2,2.71) (4,9.14) (6,18.79) (8,25.04) (10,35.23)}; 
			\addplot[blue,thick,mark=+] coordinates{(0,0) (2,2.56) (4,9.79) (6,21.22) (8,32.79) (10,41.63)}; 
			\addplot[ForestGreen,thick,mark=x] coordinates{(0,0) (2,0.81) (4,4.45) (6,10.30) (8,20.23) (10,31.33)}; 
			\addplot[cyan,thick,mark=square] coordinates{(0,0) (2,2.85) (4,8.72) (6,16.39) (8,28.01) (10,39.61)}; 
			\addplot[Gray,thick,mark=pentagon] coordinates{(0,0) (2,3.57) (4,15.27) (6,32.56) (8,38.11) (10,52.26)}; 
			\addplot[Orchid,thick,mark=diamond] coordinates{(0,0) (2,0.71) (4,7.94) (6,19.92) (8,29.27) (10,39.74)}; 
			
			\addplot[Rhodamine,thick,mark=triangle] coordinates{(0,0) (2,3.31) (4,12.49) (6,27.19) (8,46.18) (10,56.30)}; 

			\addplot[red,thick,mark=o] coordinates{(0,0) (2,3.01) (4,13.38) (6,46.84) (8,70.20) (10,87.97)}; 
			\end{axis}
			\end{tikzpicture}\\
			\begin{tikzpicture}[/pgfplots/width=0.5\linewidth,/pgfplots/height=0.288\linewidth]
			\begin{axis}
			[ymin=0,ymax=100,xmin=0,xmax=10,font=\footnotesize,
			xtick={0,2,4,6,8,10},
			xticklabels={0,2,4,6,8,10},
			title={(c) Same CNN model, different dataset},
			title style={yshift=-1.1ex},
			xlabel={{$\ell_\infty$ norm}},
			ylabel={{mean mAP Drop Rate, mDR [\%]}},
			ylabel style={yshift=-1.2ex},
			legend style={at={(0.16,0.999)},draw=none,anchor=north,legend columns=1,font=\scriptsize,fill=white,fill opacity=0.0,text opacity=1},
			legend cell align={left},
			legend entries={TCIAA*~\cite{tripletssim},AMA*\cite{bai2019},ODFA*\cite{zheng2018},GD-UAP\cite{gduap},IR-UAP\cite{lijie}, M-DI$^2$-FGSM*~\cite{M-DI-FGSM},MI-FGSM*~\cite{MI-FGSM},  TAP*~\cite{zhou2018transferable},  Ours-MUAP},
			grid=both,
			grid style=dotted,
			major grid style={white!20!black},
			minor grid style={white!70!black}]
			\addplot[LimeGreen,thick,mark=triangle] coordinates{(0,0) (2,3.45) (4,14.87) (6,28.42) (8,51.68) (10, 79.04)}; 
			\addplot[YellowOrange,thick,mark=star] coordinates{(0,0) (2,2.49) (4,10.26) (6,23.47) (8,48.10) (10,74.63)}; 
			\addplot[blue,thick,mark=+] coordinates{(0,0) (2,2.42) (4,11.13) (6,26.61) (8,53.01) (10,80.46)};  
			\addplot[ForestGreen,thick,mark=x] coordinates{(0,0) (2,1.56) (4,8.98) (6,33.01) (8,59.10) (10,79.75)};  
			\addplot[cyan,thick,mark=square] coordinates{(0,0) (2,3.26) (4,11.92) (6,27.66) (8,48.18) (10,70.47)};  
			\addplot[Gray,thick,mark=pentagon] coordinates{(0,0) (2,4.32) (4,18.10) (6,39.85) (8,51.29) (10,71.16)}; 
			\addplot[Orchid,thick,mark=diamond] coordinates{(0,0) (2,1.45) (4,12.46) (6,25.32) (8,45.19) (10,64.40)}; 
			
			\addplot[Rhodamine,thick,mark=triangle] coordinates{(0,0) (2,3.45) (4,21.75) (6,33.77) (8,58.99) (10, 72.01)}; 
			
			\addplot[red,thick,mark=o] coordinates{(0,0) (2,1.89) (4,25.01) (6,50.29) (8,73.73) (10,80.35)}; 
			
			\end{axis}
			\end{tikzpicture}&
			\begin{tikzpicture}[/pgfplots/width=0.5\linewidth,/pgfplots/height=0.288\linewidth]
			\begin{axis}
			[ymin=0,ymax=100,xmin=0,xmax=10,font=\footnotesize,
			xtick={0,2,4,6,8,10},
			xticklabels={0,2,4,6,8,10},
			title={(d) Different CNN model, different dataset},
			title style={yshift=-1.1ex},
			xlabel={{$\ell_\infty$ norm}},
			ylabel={{mean mAP Drop Rate, mDR [\%]}},
			ylabel style={yshift=-1.2ex},
			legend style={at={(0.16,0.999)},draw=none,anchor=north,legend columns=1,font=\scriptsize,fill=white,fill opacity=0.0,text opacity=1},
			legend cell align={left},
			legend entries={TCIAA*~\cite{tripletssim},AMA*\cite{bai2019},ODFA*\cite{zheng2018},GD-UAP\cite{gduap},IR-UAP\cite{lijie},M-DI$^2$-FGSM*~\cite{M-DI-FGSM},MI-FGSM*~\cite{MI-FGSM},   TAP*~\cite{zhou2018transferable},  Ours-MUAP},
			grid=both,
			grid style=dotted,
			major grid style={white!20!black},
			minor grid style={white!70!black}]
			\addplot[LimeGreen,thick,mark=triangle] coordinates{(0,0) (2,1.70) (4,4.96) (6,10.79) (8,17.75) (10,26.18)}; 
			\addplot[YellowOrange,thick,mark=star] coordinates{(0,0) (2,0.80) (4,4.05) (6,10.23) (8,19.51) (10,32.58)}; 
			\addplot[blue,thick,mark=+] coordinates{(0,0) (2,0.67) (4,4.12) (6,10.24) (8,18.80) (10,31.07)};  
			\addplot[ForestGreen,thick,mark=x] coordinates{(0,0) (2,1.08) (4,5.81) (6,13.22) (8,27.35) (10,46.95)};  
			\addplot[cyan,thick,mark=square] coordinates{(0,0) (2,0.90) (4,3.86) (6,8.50) (8,16.35) (10,28.54)};  
			\addplot[Gray,thick,mark=pentagon] coordinates{(0,0) (2,2.36) (4,8.06) (6,16.94) (8,15.57) (10,24.86)}; 
			\addplot[Orchid,thick,mark=diamond] coordinates{(0,0) (2,0.82) (4,5.36) (6,11.39) (8,12.14) (10,19.17)}; 
			
			\addplot[Rhodamine,thick,mark=triangle] coordinates{(0,0) (2,2.76) (4,10.74) (6,18.27) (8,33.93) (10,44.16)}; 
			
			\addplot[red,thick,mark=o] coordinates{(0,0) (2,0.63) (4,14.68) (6,38.13) (8,66.32) (10,83.56)};  
			\end{axis}
			\end{tikzpicture}\\
		\end{tabular}
		\caption{\textbf{Comparison with state-of-the-art attack methods.} \emph{State-of-the-art methods work well in the white-box attack scenario, but degenerate significantly in the black-box case where the proposed method outperforms the others by a large margin.} We compare our proposed method with TCIAA~\cite{tripletssim}, AMA~\cite{bai2019}, ODFA~\cite{zheng2018}, GD-UAP~\cite{gduap}, IR-UAP~\cite{lijie}, M-DI$^2$-FGSM~\cite{M-DI-FGSM}, MI-FGSM~\cite{MI-FGSM},  and TAP~\cite{zhou2018transferable}. (a) shows the attack performances in white-box scenario where UAP is trained and tested on the same model employing the same dataset. (b-d) show performances in transfer attack or black-box attack scenario where UAP is evaluated across CNN architectures or datasets. The asterisk (*) denotes methods that are adapted from original ones to the `universal' setting.} 
		\label{fig:compare}
	\end{figure*}

	In this section, we elaborate the evaluation metrics, the datasets, and the experimental settings.
	{\flushleft\textbf{Evaluation metric.}} 
	Following~\cite{lijie}, we compute the mean \emph{mAP} Drop Rate ($mDR$) to evaluate the attack performance quantitatively:
	\begin{equation}
	mDR(q,u) = \frac{mAP(q)-mAP(q+u)}{mAP(q)},
	\label{eq:mdr}
	\end{equation}
	where $mAP(q)$ denotes the mean average precision of $q$, and $u$ refers to our UAP vector. A positive $mDR$ means the attack is successful, and higher the value, better the attack. $mAP$ metric in Eq.\ref{eq:mdr} can be replaced by other Re-ID metrics such as Rank-1, and the mean Rank-1 Drop Rate~($RDR$) is obtained in the same way.
	
	{\flushleft\textbf{Datasets.}} We perform experiments on three widely used image-based Re-ID datasets and one video-based dataset, including Market-1501~\cite{market}, DukeMTMC-reID~\cite{duke}, MSMT17~\cite{msmt17}, and MARS~\cite{mars}. Market-1501 contains 32668 pictures corresponding to 1501 identities. 12936 images are used for training and the rest used for testing. DukeMTMC-reID contains 36411 images of 1404 individuals, of which 702 identities are used for training and the rest 702 identities for testing. MSMT17 contains 126441 images of 4101 identities. 32621 images of 1041 identities are used for training and the rest used for testing. MARS is the largest video-based Re-ID dataset with around 20000 tracklets corresponding to 1261 identities. For image-based Re-ID models, We perform attack by adding UAP on query images and aim to cause lower rank of gallery images corresponding to the same identity. For video-based ones, we add UAP on every frame of the query clips. 
	{\flushleft\textbf{Implementation details.}} To evaluate the UAP transferability to different CNN architectures, we train models with five different backbones following the effective training method proposed in BoT~\cite{bag-of-tricks}, including ResNet50~\cite{resnet50}, DenseNet121~\cite{densenet}, VGG16~\cite{vgg}, SENet154~\cite{senet}, and ShuffleNet~\cite{shufflenet}. More details of the performances of these models before and after attack are reported in \textit{Appendix~\ref{attacksota}}.
	We train UAP with 800 images randomly selected from the corresponding training set. The input images are resized to $256\times128$, and we set $\lambda=10$, $\eta=0.4$, the learning rate $\alpha=0.25$, the maximum number of epochs $T=100$, and the training batchsize is set to 32. If not specified, $\ell_2$ form of MI regularization is adopted in this paper. 
	
	\section{Results}
	\label{sec:results}
	In this section, we perform extensive experiments to validate the effectiveness of our proposed method. 1)~We compare the attack performance of our proposed methods with \emph{eight} state of the arts in different scenarios for sufficient comparison. As most of the methods are image-specific, we build the baselines by adapting these methods\cite{bai2019,zheng2018,tripletssim,zhou2018transferable,MI-FGSM,M-DI-FGSM} to the {universal} setting and compare our method with them.\footnote{ To compare with AMA~\cite{bai2019}, OFDA~\cite{zheng2018}, TCIAA~\cite{tripletssim}, and TAP~\cite{zhou2018transferable}, as these methods focus on objective functions, we generalize them to corresponding `universal' versions by replacing  our proposed objective function in Eq.~\ref{eq:all} with theirs. For  MI-FGSM~\cite{MI-FGSM} and M-DI$^2$-FGSM~\cite{M-DI-FGSM}, we adapt in a similar way by replacing our proposed objective as their classification objective. } 2)~To further demonstrate the effectiveness of each component of our method, we perform extensive ablation study in various scenarios. 3)~We evaluate the attack transferability of UAP to models with different CNN architectures or to different datasets. 4)~We test different forms of the model-insensitive regularization with $\ell_1$ and $\ell_2$ norm respectively. 
	\subsection{Comparison with State of The Arts}
	\label{subsec:comp_sota}
	We compare our proposed method with state-of-the-art attack methods in the `universal' setting, including TCIAA~\cite{tripletssim}, AMA~\cite{bai2019}, ODFA~\cite{zheng2018}, GD-UAP~\cite{gduap}, IR-UAP~\cite{lijie}, M-DI$^2$-FGSM~\cite{M-DI-FGSM}, MI-FGSM~\cite{MI-FGSM},  and TAP~\cite{zhou2018transferable}. The experimental results are reported in Fig.~\ref{fig:compare}. We test the transferability of different methods in two cases: cross-CNN model and cross-dataset. In the cross-model protocol, UAPs are trained using ResNet50 while tested on DenseNet121. And in the cross-dataset setting, UAPs are trained on the MSMT17 dataset while tested on Market1501.

	Overall, the proposed method significantly outperforms other methods in the \emph{black-box} scenario (Fig.~\ref{fig:compare}~(b)-(d)) and also achieves the best performance as state of the arts in \emph{white-box} scenario (Fig.~\ref{fig:compare}~(a)). 

	Other highlights in Fig.~\ref{fig:compare} can be summarized as follows: 

	\begin{itemize}
		\item \textit{Classification-based vs. similarity-based.} Most of the methods perform well in the white-box scenario, where similarity-based methods outperform classification-based methods in general, \eg ours-MUAP vs. MI-FGSM, and ODFA vs. TAP. As can be seen, the proposed MUAP achieves the best performance in various scenarios.

		\item Our proposed method outperforms all the other methods by a large margin in black-box scenario. As shown in (b)-(d), the performances of most methods degenerate significantly when testing on a different CNN model or using a different test set. However, the proposed MUAP shows high transferability in cross-model and cross-dataset scenarios. 
	\end{itemize}

	\subsection{Ablation Study}
	\begin{table*}[!th]
		\renewcommand\arraystretch{1.2}
		\centering\scriptsize
		\caption{\textbf{Cross-model attack results.} Our MUAP attack performance evaluation within the same dataset using $\|u\|_{\infty} \le 10$.}\smallskip
		\resizebox{1.0\textwidth}{!}{
			\renewcommand{\tabcolsep}{0.0pt}
			\begin{tabular}[c]{| c | c || C{0.4cm} | C{1.5cm} | C{1.5cm} | C{1.5cm} | C{1.5cm} | C{1.5cm} || C{0.4cm} | C{1.5cm} | C{1.5cm} | C{1.5cm} | C{1.5cm} | C{1.5cm} |}
				\hline
				\multicolumn{2}{|c||}{mean mAP Drop Rate} &  &
				\multicolumn{5}{c||}{\textbf{Test: DukeMTMC-reID}} & &
				\multicolumn{5}{c|}{\textbf{Test: Market-1501}} \\
				\cline{4-8} \cline{10-14}
				\multicolumn{2}{|c||}{(mDR)} & & ResNet50 & DenseNet121 & VGG16 &  SENet154 & ShuffleNet & &  ResNet50 & DenseNet121 & VGG16 & SENet154 & ShuffleNet \\
				\hline 
				\hline
				\multirow{3}{*}{ResNet50} & $\mathcal{L}_{Base}$ & \multirow{15}{*}{\rotatebox{90}{\textbf{Train: DukeMTMC-reID}}} & \cellcolor{Gray!30}84.38\% & 29.75\% & 38.15\% & 30.53\% & 30.96\% & \multirow{15}{*}{\rotatebox{90}{\textbf{Train: Market-1501}}} & \cellcolor{Gray!30}89.78\%	& 38.24\%& 34.91\% & 34.92\%  & 60.09\%\\
				 	 	 	 	  	 	 	 	 
				& $\mathcal{L}_{AP}$ & &  \cellcolor{Gray!30}\textbf{98.49}\% & 33.61\% & 49.40\% & 21.89\% & 30.31\% & & \cellcolor{Gray!30}\textbf{99.39\%} & 40.47\% & 46.11\% & 28.40\%& 55.11\%\\
				 	 	 	 	 	 	 	 	 
				& $+\mathcal{L}_{MI}$ & & \cellcolor{Gray!30}92.14\%	& \textbf{75.96\%}& 	\textbf{81.51\%}& 	\textbf{71.28\%}& 	\textbf{71.68\%}&&		\cellcolor{Gray!30}93.16	\%&\textbf{87.97\%}&	\textbf{92.17\%}&	\textbf{97.10\%}&	\textbf{87.04\%}\\
				\cline{1-2} \cline{4-8} \cline{10-14}	 	 	 
	
				\multirow{3}{*}{DenseNet121} & $\mathcal{L}_{Base}$ & & 62.93\%&	\cellcolor{Gray!30}81.96\%& 52.02\% & 36.46\%&	46.11\%&&	43.92\%&	\cellcolor{Gray!30}83.53\% & 41.54\%&	27.87\%&	54.04\% \\
				 	 	 	 	 
				& $\mathcal{L}_{AP}$ & & \textbf{89.89\%}&	\cellcolor{Gray!30}\textbf{98.87\%}&	73.15\%&	54.83\%&	71.12\%&& 66.11\%&	\cellcolor{Gray!30}\textbf{99.36\%}&	73.00\%&	32.44\%&	79.68\%\\
				
				& $+\mathcal{L}_{MI}$ & & 84.55\%&	\cellcolor{Gray!30}96.44\%&	\textbf{81.00\%}&	\textbf{74.58\%}&	\textbf{84.26\%}&&		\textbf{84.69\%}&	\cellcolor{Gray!30}98.48\%&	\textbf{89.37\%}&	\textbf{85.86\%}&	\textbf{86.25\%}\\
				\cline{1-2} \cline{4-8} \cline{10-14}
				
				\multirow{2}{*}{VGG16} & $\mathcal{L}_{Base}$ & & 51.41\%&	28.77\%&	\cellcolor{Gray!30}89.03\%&	22.99\%&	27.9\%&&		19.96\%&	17.03\%&	\cellcolor{Gray!30}95.78\% &	13.52\%&	37.38\% \\ 
				
				& $\mathcal{L}_{AP}$ & & 65.91\%&	38.34\%&	\cellcolor{Gray!30}\textbf{98.93\%}&	25.28\%&	35.77\%&&		24.44\%&	23.17\%&	\cellcolor{Gray!30}\textbf{99.42\%}&	14.45\%&	42.75\% \\
				 	 	 	 	 
				& $+\mathcal{L}_{MI}$ & & \textbf{81.27\%}&	\textbf{82.00\%}&	\cellcolor{Gray!30}96.87\%&	\textbf{81.15\%}&	\textbf{82.45\%}&&		\textbf{86.43\%}&	\textbf{83.90\%}&	\cellcolor{Gray!30}98.89\%&	\textbf{96.74\%}&	\textbf{89.03\%} \\
				\cline{1-2} \cline{4-8} \cline{10-14}
				
				\multirow{2}{*}{SENet154} &$\mathcal{L}_{Base}$&& 47.49\%&	45.62\%&	55.01\%&	\cellcolor{Gray!30}77.47\%&	47.49\%&&		28.97\%&	26.29\%&	35.15\%&	\cellcolor{Gray!30}{85.69\%}& 46.26\%\\
				 	 	 	 	 
				&$\mathcal{L}_{AP}$&& 76.18\%&	72.49\%&	83.32\%&	\cellcolor{Gray!30}\textbf{97.27\%}&	74.75\%&&	59.82\%&	51.33\%&	58.05\%&	\cellcolor{Gray!30}\textbf{98.95\%}&	74.18\%\\
				 	 	 	 	  	 	 	 	 
				& $+\mathcal{L}_{MI}$ & & \textbf{79.27\%}&	\textbf{78.66\%}&	\textbf{89.78\%}&	\cellcolor{Gray!30}93.60\%&	\textbf{83.10\%}&& \textbf{81.68\%}&	\textbf{77.80\%}&	\textbf{90.93\%}&	\cellcolor{Gray!30}97.82\%&	\textbf{86.65\%}\\
				\cline{1-2} \cline{4-8} \cline{10-14}
				 	 	 	 	 
				\multirow{2}{*}{ShuffleNet} & $\mathcal{L}_{Base}$ & &  16.80\%&	13.22\%&	18.06\%&	14.44\%&	\cellcolor{Gray!30}85.70\%&&	13.48\%&	12.28\%&	13.36\%&	12.88\%&	\cellcolor{Gray!30}{95.52\%}\\
				
				& $\mathcal{L}_{AP}$ & &  13.78\%&	11.95\%&	13.21\%&	10.33\%&	\cellcolor{Gray!30}\textbf{98.94\%}&&		24.46\%&	24.31\%&	22.20\%&	24.42\%&	\cellcolor{Gray!30}\textbf{99.75\%}\\
				 		 
				& $+\mathcal{L}_{MI}$ & & \textbf{78.30\%}&	\textbf{80.10\%}&	\textbf{84.18\%}&	\textbf{84.93\%}&	\cellcolor{Gray!30}93.44\%&&		\textbf{71.27\%}&	\textbf{74.02\%}&	\textbf{81.72\%}&	\textbf{89.18\%}&	\cellcolor{Gray!30}98.86\%\\
				\hline
			\end{tabular}
		}
		\label{single-dataset-l2}
	\end{table*}
	\begin{table*}[!th]
		\renewcommand\arraystretch{1.2}
		\centering\scriptsize
		\caption{\textbf{Cross-model attack results in terms of mean Rank-1 drop rate.} }
		\resizebox{1.0\textwidth}{!}{
			\renewcommand{\tabcolsep}{0.0pt}
			\begin{tabular}[c]{| c | c || C{0.4cm} | C{1.5cm} | C{1.5cm} | C{1.5cm} | C{1.5cm} | C{1.5cm} || C{0.4cm} | C{1.5cm} | C{1.5cm} | C{1.5cm} | C{1.5cm} | C{1.5cm} |}
				\hline
				\multicolumn{2}{|c||}{mean Rank-1 Drop Rate} &  &
				\multicolumn{5}{c||}{\textbf{Test: DukeMTMC-reID}} & &
				\multicolumn{5}{c|}{\textbf{Test: Market-1501}} \\
				\cline{4-8} \cline{10-14}
				\multicolumn{2}{|c||}{(RDR)} & & ResNet50 & DenseNet121 & VGG16 &  SENet154 & ShuffleNet & &  ResNet50 & DenseNet121 & VGG16 & SENet154 & ShuffleNet \\
				\hline 
				\hline
				\multirow{3}{*}{ResNet50} & $\mathcal{L}_{Base}$ & \multirow{15}{*}{\rotatebox{90}{\textbf{Train: DukeMTMC-reID}}} & \cellcolor{Gray!30}83.59\%	& 21.90\% &	33.35\%	 & 24.34\% & 22.66\% & \multirow{15}{*}{\rotatebox{90}{\textbf{Train: Market-1501}}} & \cellcolor{Gray!30}89.02\%&	27.24\%	&27.80\%&	26.12\%&	51.09\% \\
				 	 	 	 	  	 	 	 	 
				& $\mathcal{L}_{AP}$ & &  \cellcolor{Gray!30}\textbf{98.71\%} &25.48\% & 43.61\% &15.62\%	& 23.55\% & & \cellcolor{Gray!30}\textbf{99.84\%
} & 30.29\% & 38.75\% & 18.31\%	& 44.39\% \\
				 	 	 	 	 	 	 	 	 
				& $+\mathcal{L}_{MI}$ & & \cellcolor{Gray!30}91.90\%
	& \textbf{72.19\%}& 	\textbf{78.78\%}& 	\textbf{66.61\%}& 	\textbf{69.43\%}&&		\cellcolor{Gray!30}93.42	\%&\textbf{87.65\%}&	\textbf{92.43\%}&	\textbf{97.57\%}&	\textbf{86.12\%}\\
				\cline{1-2} \cline{4-8} \cline{10-14}	 	 	 
	
				\multirow{3}{*}{DenseNet121} & $\mathcal{L}_{Base}$ & & 58.51\%
&	\cellcolor{Gray!30}80.09\% &	46.31\% &	30.69\%	 & 38.80\%
&&	35.24\%&	\cellcolor{Gray!30}78.41\%&	34.14\%&	18.64\%&	43.83\% \\
				 	 	 	 	 
				& $\mathcal{L}_{AP}$ & & \textbf{90.04\%}&	\cellcolor{Gray!30}\textbf{99.42\%}& 70.18\% & 49.56\%	& 67.05\%
&& 60.93\% &	\cellcolor{Gray!30}\textbf{99.94\%} & 72.04\%	& 21.04\% & 76.43\%\\
				
				& $+\mathcal{L}_{MI}$ & & 83.64\%&	\cellcolor{Gray!30}96.52\%&	\textbf{79.71\%}&	\textbf{69.48\%
}&	\textbf{83.36\%}&&		\textbf{84.88\%}&	\cellcolor{Gray!30}99.84\%&	\textbf{89.57\%}&	\textbf{84.48\%}&	\textbf{85.56\%}\\
				\cline{1-2} \cline{4-8} \cline{10-14}
				
				\multirow{2}{*}{VGG16} & $\mathcal{L}_{Base}$ & & 44.73\%&	21.48\%&	\cellcolor{Gray!30}87.82\%&	17.00\%&	21.17\%&&		11.20\%&	9.66\%&	\cellcolor{Gray!30}96.94\% &	7.48\%&	27.12\% \\ 
				
				& $\mathcal{L}_{AP}$ & & 60.63\% & 29.64\%
&	\cellcolor{Gray!30}\textbf{99.56\%}&	18.99\%&	28.35\%&&		15.62\%&	14.05\%&	\cellcolor{Gray!30}\textbf{99.84\%}&	8.01\%&	32.53\% \\
				 	 	 	 	 
				& $+\mathcal{L}_{MI}$ & & \textbf{78.22\%}&	\textbf{78.41\%}&	\cellcolor{Gray!30}97.24\%&	\textbf{78.75\%}&	\textbf{80.76\%}&&		\textbf{85.35\%}&	\textbf{81.26\%}&	\cellcolor{Gray!30}99.67\%&	\textbf{97.04\%}&	\textbf{88.60\%} \\
				\cline{1-2} \cline{4-8} \cline{10-14}
				
				\multirow{2}{*}{SENet154} &$\mathcal{L}_{Base}$&& 40.40\%&	38.44\%&	50.55\%&	\cellcolor{Gray!30}73.34\% &	40.02\%&&		20.88\%&	16.49\%&	29.51\%&	\cellcolor{Gray!30}{82.98\%}& 35.86\%\\
				 	 	 	 	 
				&$\mathcal{L}_{AP}$&& 72.70\%&	69.09\%&	80.98\%&	\cellcolor{Gray!30}\textbf{97.24\%}&	72.47\%&&	53.91\%&	42.29\%&	52.66\%&	\cellcolor{Gray!30}\textbf{99.37\%}&	69.18\%\\
				 	 	 	 	  	 	 	 	 
				& $+\mathcal{L}_{MI}$ & & \textbf{76.73\%}&	\textbf{75.20\%}&	\textbf{88.86\%}&	\cellcolor{Gray!30}92.72\%&	\textbf{81.48\%}&& \textbf{76.05\%}&	\textbf{71.74\%}&	\textbf{87.30\%}&	\cellcolor{Gray!30}99.07\%&	\textbf{84.41\%}\\
				\cline{1-2} \cline{4-8} \cline{10-14}
				 	 	 	 	 
				\multirow{2}{*}{ShuffleNet} & $\mathcal{L}_{Base}$ & &  10.83\%	&7.53\%&	13.67\%&	10.21\% &
	\cellcolor{Gray!30}82.64\%&&	6.64\%&	6.23\%&	9.41\%&	6.55\%&	\cellcolor{Gray!30}{95.52\%}\\
				
				& $\mathcal{L}_{AP}$ & &  9.08\%&	6.53\%&	10.42\%&	6.62\%&
	\cellcolor{Gray!30}\textbf{99.00\%}&&		15.59\%&	16.01\%&	16.48\%&	15.65\%&	\cellcolor{Gray!30}\textbf{99.99\%}\\
				 		 
				& $+\mathcal{L}_{MI}$ & & \textbf{76.21\%}&	\textbf{77.51\%}&	\textbf{81.75\%}&	\textbf{82.89\%}&	\cellcolor{Gray!30}93.31\%&&		\textbf{66.94\%}&	\textbf{71.48\%}&	\textbf{80.10\%}&	\textbf{88.40\%}&	\cellcolor{Gray!30}99.97\%\\
				\hline
			\end{tabular}
		}
		\label{single-dataset-rank1}
	\end{table*}

	\begin{table*}[!th]
		\centering\scriptsize
		\caption{\textbf{Cross-dataset attack results.} Our MUAP attack performance evaluation across different datasets using $\|u\|_{\infty} \le 10$.}\smallskip
		\resizebox{1.0\textwidth}{!}{
			\renewcommand{\arraystretch}{1.2}
			\renewcommand{\tabcolsep}{0.0pt}
			\begin{tabular}[c]{| C{0.3cm} | c | c || C{1.6cm} | C{1.6cm} | C{1.6cm} | C{1.6cm} | C{1.6cm} || C{1.6cm} | C{1.6cm} | C{1.6cm} | C{1.6cm} | C{1.6cm} |}
				\hline
				\multicolumn{3}{|c||}{mean mAP Drop Rate} & 
				\multicolumn{5}{c||}{\textbf{Test: DukeMTMC-reID}} &
				\multicolumn{5}{c|}{\textbf{Test: Market-1501}} \\
				\cline{4-13}
				\multicolumn{3}{|c||}{(mDR)} & ResNet50 & DenseNet121 & VGG16 & SENet154 & ShuffleNet &  ResNet50 & DenseNet121 & VGG16 & SENet154 & ShuffleNet \\
				\hline
				\hline
				\multirow{15}{*}{\rotatebox{90}{\textbf{Train: MSMT17}}}
			&\multirow{2}{*}{ResNet50}
			& $\mathcal{L}_{Base}$ & \cellcolor{Gray!30}56.4\% & 25.25\% & 32.65\% & 25.04\% & 28.36\% & \cellcolor{Gray!30}{52.34\%}& 27.89\%& 33.34 \%&21.24\%& 47.69\%\\
			&& $\mathcal{L}_{AP}$ & \cellcolor{Gray!30}70.22\% & 17.80\% & 26.05\% & 10.83\% & 36.46\% & \cellcolor{Gray!30}\textbf{83.70\%}& 29.57\%& 38.83 \%&21.43\%& 55.67\%\\
			&& $+\mathcal{L}_{MI}$ &  \cellcolor{Gray!30}\textbf{83.36\%}&	\textbf{83.01\%}&	\textbf{88.12\%}&	\textbf{79.34\%}&	\textbf{67.45\%}&		\cellcolor{Gray!30}80.35\%&	\textbf{82.93\%}&	\textbf{89.38\%}&	\textbf{85.5\%}&	\textbf{68.38\%}\\
			\cline{2-13}
			&\multirow{2}{*}{DenseNet121} & $\mathcal{L}_{Base}$ & 39.64\%&	\cellcolor{Gray!30}45.33\%&	39.67\%&	19.13\%&	30.80\%&		30.62\%&	\cellcolor{Gray!30}43.61\%&	32.73\%&	18.18\%&	43.66\%\\
			
			&& $\mathcal{L}_{AP}$ & 33.84\%&	\cellcolor{Gray!30}34.41\%&	28.39\%&	15.05\%&	32.50\%&		34.76\%&	\cellcolor{Gray!30}54.37\%&	35.65\%&	22.4\%&	49.97\%\\
			
			&& $+\mathcal{L}_{MI}$ & \textbf{82.37\%}&	\cellcolor{Gray!30}\textbf{91.24\%}&	\textbf{89.94\%}&	\textbf{87.55\%}&	\textbf{81.67\%}&		\textbf{75.99\%}&	\cellcolor{Gray!30}\textbf{92.64\%}&\textbf{	92.43\%}&	\textbf{91.21\%}&	\textbf{85.66\%}\\ 
			\cline{2-13}
			&\multirow{2}{*}{VGG16} & $\mathcal{L}_{Base}$ & 25.26\%&	14.20\%&	\cellcolor{Gray!30}39.99\%&	10.28\%&	26.37\%&		19.23\%&	16.89\%&	\cellcolor{Gray!30}57.09\%&	11.92\%&	38.13\%\\
			
			&& $\mathcal{L}_{AP}$ & 39.83\%&	20.71\%&	\cellcolor{Gray!30}47.46\%&	11.01\%&	38.89\%&		23.15\%&	20.51\%&	\cellcolor{Gray!30}60.06\%&	15.72\%&	45.81\%\\
			&& $+\mathcal{L}_{MI}$ & \textbf{78.50\%}&	\textbf{85.87\%}&	\cellcolor{Gray!30}\textbf{89.02\%}&	\textbf{85.12\%}&	\textbf{72.32\%}&		\textbf{72.25\%}&	\textbf{81.51\%}&	\cellcolor{Gray!30}\textbf{93.34\%}&	\textbf{93.06\%}&	\textbf{73.48 \%}\\
			\cline{2-13}
			&\multirow{2}{*}{SENet154} 
			& $\mathcal{L}_{Base}$ & 55.86\%&	47.53\%& 59.40\%&\cellcolor{Gray!30}	\cellcolor{Gray!30}55.07\%&	48.92\%&		28.20\%&25.26\%&	47.94\%&	\cellcolor{Gray!30}42.21\%&	57.93\%\\
			
			&& $\mathcal{L}_{AP}$ & 70.66\%&	71.75\%&	77.83\%&\cellcolor{Gray!30}	\cellcolor{Gray!30}80.67\%&	67.25\%&		58.32\%&	65.24\%&	77.97\%&	\cellcolor{Gray!30}88.28\%&	79.13\%\\
			&& $+\mathcal{L}_{MI}$&\textbf{76.61\%}&	\textbf{76.81\%}&	\textbf{85.9\%}&	\cellcolor{Gray!30}\textbf{91.35\%}&	\textbf{76.51\%}&		\textbf{73.82\%}&	\textbf{73.09\%}&	\textbf{88.63\%}&	\cellcolor{Gray!30}\textbf{95.74\%}&	\textbf{79.20\%}\\
			\cline{2-13}
			&\multirow{2}{*}{ShuffleNet} 
			& $\mathcal{L}_{Base}$ & 10.37\%&	7.25\%&	12.09\%&	6.55\%&	\cellcolor{Gray!30}{64.50\%}&		9.23\%&	9.24\%&	12.91\%&	9.31\%&	\cellcolor{Gray!30}{71.03\%}\\
			&& $\mathcal{L}_{AP}$ & 11.76\%&	6.97\%&	10.44\%&	7.41\%&	\cellcolor{Gray!30}\textbf{89.65\%}& 12.26\%&	12.47\%&	16.33\%&	17.98\%&	\cellcolor{Gray!30}\textbf{97.73\%}\\
			&& $+\mathcal{L}_{MI}$ &  \textbf{80.35\%}&	\textbf{82.21\%}&	\textbf{89.69\%}&	\textbf{94.35\%}&	\cellcolor{Gray!30}88.09\%&		\textbf{81.77\%}&	\textbf{80.31\%}&	\textbf{89.09\%}&	\textbf{96.42\%}&	\cellcolor{Gray!30}90.24\%\\
			\hline
			\end{tabular}
		}
		\label{cross-dataset-l2}
	\end{table*}
	
	\begin{table*}[!th]
		\centering\scriptsize
		\caption{\textbf{Cross-dataset attack results in terms of mean Rank-1 drop rate.} }\smallskip 
		\resizebox{1.0\textwidth}{!}{
			\renewcommand{\arraystretch}{1.2}
			\renewcommand{\tabcolsep}{0.0pt}
			\begin{tabular}[c]{| C{0.4cm} | c | c || C{1.6cm} | C{1.6cm} | C{1.6cm} | C{1.6cm} | C{1.6cm} || C{1.6cm} | C{1.6cm} | C{1.6cm} | C{1.6cm} | C{1.6cm} |}
				\hline
				\multicolumn{3}{|c||}{mean Rank-1 Drop Rate} & 
				\multicolumn{5}{c||}{\textbf{Test: DukeMTMC-reID}} &
				\multicolumn{5}{c|}{\textbf{Test: Market-1501}} \\
				\cline{4-13}
				\multicolumn{3}{|c||}{(RDR)} & ResNet50 & DenseNet121 & VGG16 & SENet154 & ShuffleNet &  ResNet50 & DenseNet121 & VGG16 & SENet154 & ShuffleNet \\
				\hline
				\hline
				\multirow{15}{*}{\rotatebox{90}{\textbf{Train: MSMT17}}}
			&\multirow{2}{*}{ResNet50}
			& $\mathcal{L}_{Base}$ & \cellcolor{Gray!30}50.36\%	&17.85\%&	28.23\%& 19.87\%	& 20.84\%
 & \cellcolor{Gray!30}{44.57\%}& 16.94\% & 26.97\%	& 13.13\%& 	37.84\%\\
			&& $\mathcal{L}_{AP}$ & \cellcolor{Gray!30}66.05\%&	11.95\%&	20.84\%&	7.12\%&	27.14\%
 & \cellcolor{Gray!30}\textbf{84.94\%}& 19.19\%& 31.05\%&	13.43\%&	45.38\% \\
			&& $+\mathcal{L}_{MI}$ &  \cellcolor{Gray!30}\textbf{79.41\%}&	\textbf{77.41\%}&	\textbf{84.56\%}&	\textbf{75.22\%}&	\textbf{63.29\%}&		\cellcolor{Gray!30}74.82\%&	\textbf{75.07\%}&	\textbf{85.89\%}&	\textbf{81.19\%}&	\textbf{63.28\%}\\
			\cline{2-13}
			&\multirow{2}{*}{DenseNet121} & $\mathcal{L}_{Base}$ & 32.20\%&	\cellcolor{Gray!30}37.65\%&	33.08\%&	13.58\%&	22.66\%&		21.92\%&	\cellcolor{Gray!30}34.33\%&	26.91\%&	10.87\%&	33.06\%\\
			
			&& $\mathcal{L}_{AP}$ & 25.90\%&	\cellcolor{Gray!30}25.91\%&	23.10\%&10.76\%&24.10\% &	23.94\%&	\cellcolor{Gray!30}44.56\%&	28.35\%&	13.39\%&	39.81\%\\
			
			&& $+\mathcal{L}_{MI}$ & \textbf{81.22\%}&	\cellcolor{Gray!30}\textbf{90.57\%}&	\textbf{88.92\%}&	\textbf{86.04\%}&	\textbf{78.66\%}&		\textbf{73.33\%}&	\cellcolor{Gray!30}\textbf{93.65\%}&\textbf{	93.52\%}&	\textbf{91.03\%}&	\textbf{84.54\%}\\ 
			\cline{2-13}
			&\multirow{2}{*}{VGG16} & $\mathcal{L}_{Base}$ & 17.44\%&	8.26\%&	\cellcolor{Gray!30}33.13\%&	6.79\%&	18.51\%&		11.13\%&	9.88\%&	\cellcolor{Gray!30}51.97\%&	6.21\%&	27.52\%\\
			
			&& $\mathcal{L}_{AP}$ & 32.76\%&	11.84\%&	\cellcolor{Gray!30}39.64\%&	7.18\%&	30.73\%&		14.17\%&	12.16\%&	\cellcolor{Gray!30}53.85\%&	9.14\%&	35.69\%\\
			&& $+\mathcal{L}_{MI}$ & \textbf{76.68\%}&	\textbf{84.68\%}&	\cellcolor{Gray!30}\textbf{87.54\%}&	\textbf{83.55\%}&	\textbf{68.10\%}&		\textbf{67.76\%}&	\textbf{80.05\%}&	\cellcolor{Gray!30}\textbf{94.37\%}&	\textbf{93.29\%}&	\textbf{67.37 \%}\\
			\cline{2-13}
			&\multirow{2}{*}{SENet154} 
			& $\mathcal{L}_{Base}$ & 49.33\%&	40.12\%&	53.58\%
&\cellcolor{Gray!30}	\cellcolor{Gray!30}48.95\%&	42.84\%&		18.91\%&13.67\%&	40.89\%&	\cellcolor{Gray!30}32.73\%&	48.75\%\\
			
			&& $\mathcal{L}_{AP}$ & 66.46\%&	68.03\%&	76.13\%&\cellcolor{Gray!30}	\cellcolor{Gray!30}78.53\%&	62.41\%&		52.83\%&	56.85\%&	75.49\%&	\cellcolor{Gray!30}86.71\%&	74.85\%\\
			&& $+\mathcal{L}_{MI}$&\textbf{73.43\%}&	\textbf{73.62\%}&	\textbf{85.06\%}&	\cellcolor{Gray!30}\textbf{91.28\%}&	\textbf{72.69\%}&		\textbf{69.94\%}&	\textbf{68.34\%}&	\textbf{88.42\%}&	\cellcolor{Gray!30}\textbf{95.38\%}&	\textbf{75.51\%}\\
			\cline{2-13}
			&\multirow{2}{*}{ShuffleNet} 
			& $\mathcal{L}_{Base}$ & 7.12\%&	3.79\%&	8.00\%&	3.86\%&	\cellcolor{Gray!30}{57.32\%}&		4.36\%&	4.69\%&	9.37\%&	5.28\%&	\cellcolor{Gray!30}{62.59\%}\\
			&& $\mathcal{L}_{AP}$ & 8.61\%& 3.58\%&	6.95\%&	4.20\%&	\cellcolor{Gray!30}\textbf{88.22\%}& 5.69\%&	7.00\%&	11.35\%&	10.57\%&	\cellcolor{Gray!30}\textbf{98.29\%}\\
			&& $+\mathcal{L}_{MI}$ &  \textbf{77.76\%}&	\textbf{79.46\%}&	\textbf{89.36\%}&	\textbf{94.32\%}&	\cellcolor{Gray!30}86.95\%&		\textbf{80.86\%}&	\textbf{79.40\%}&	\textbf{88.78\%}&	\textbf{96.81\%}&	\cellcolor{Gray!30}90.87\%\\
			\hline
			\end{tabular}
		}
		\label{cross-dataset-rank1}
	\end{table*}

	To evaluate the performance of each component of our method, we perform extensive ablation study. In every three rows in Tables~\ref{single-dataset-l2}~,~\ref{single-dataset-rank1},~\ref{cross-dataset-l2},~\ref{cross-dataset-rank1}, we compare the baseline, the one with $\mathcal{L}_{AP}$ and with $\mathcal{L}_{AP}$ and $\mathcal{L}_{MI}$ respectively. Details are reported as follows. 
	
	{\flushleft\textbf{Cross CNN architecture attack within the same dataset.}}
	We report the $mDR$ on DukeMTMC-reID and Market-1501 using $\{\gamma=\infty, \epsilon=10, \lambda=10\}$ in Table~\ref{single-dataset-l2}. The corresponding $RDR$ results are reported in Table~\ref{single-dataset-rank1}. For every dataset, each element in the table represents the $mDR$ (or $RDR$) when UAP is trained on the model of the row to attack the model of the column. The results on the diagonal of the table marked in gray represent performances in the white-box scenario. As can be seen, the proposed list-wise attack~($\mathcal{L}_{AP}$) consistently outperforms the traditional classification-based attack method~($\mathcal{L}_{Base}$) in terms of $mDR$ and $RDR$~(Table~\ref{single-dataset-l2}, Table~\ref{single-dataset-rank1}), which validate the superiority of the proposed list-wise attack. For example, the list-wise attack outperforms the classification-based attack by $14.11\%$ in terms of $mDR$ when trained and tested on ResNet50 on DukeMTMC-reID. The proposed method achieves $98\%+$ $mDR$ in most cases using $\mathcal{L}_{AP}$, which means that the ranking list is almost entirely disrupted under attack. However, the transferability of UAP to other CNN architectures is unsatisfying. For instance, the $mDR$ is only $12.88\%$ when using the UAP trained on ShuffleNet to attack SENet154 on Market-1501.    
	On the other hand, our method with model-insensitive regularization~($+\mathcal{L}_{MI}$) significantly improves the transferability to other CNN architectures with minor performance loss under the same backbone. Specifically, we improve the attack performance from $29.75\%$ to $75.96\%$ in terms of $mDR$ on DukeMTMC-reID when using the UAP trained on ResNet50 to attack DenseNet121 as shown in Table~\ref{single-dataset-l2}. 
		\begin{table*}[!th]
  			\centering\scriptsize
  			\caption{\textbf{Accuracy (mAP) of BoT models before and after attack.} (R) denotes ResNet50 for short; (D) denotes DenseNet121; (V) denotes VGG16; (SE) denotes SENet154; (Sh) denotes ShuffleNet. Bot(D) denotes the BoT model trained with DenseNet121 backbone. The corresponding \textit{mean mAP Drop Rate (mDR)} is also presented in parentheses.}\smallskip
  			\resizebox{1.0\textwidth}{!}{
  				\renewcommand{\arraystretch}{1.6}
  				\renewcommand{\tabcolsep}{0.0pt}
  				\begin{tabular}[c]{|C{1.5cm} || C{1.4cm} | C{1.4cm} | C{1.4cm} | C{1.4cm} | C{1.4cm} || C{1.4cm} | C{1.4cm} | C{1.4cm} | C{1.4cm} |C{1.4cm} |}
  					\hline
  					\multicolumn{1}{|c||}{\multirow{2}{*}{\small mAP}} & 
  					\multicolumn{5}{c||}{\textbf{DukeMTMC-reID}} &
  					\multicolumn{5}{c|}{\textbf{Market-1501}} \\
  					\cline{2-11}
  					
  					&BoT(R) & BoT(D) & BoT(V) & BoT(SE) & BoT(Sh)& BoT(R) & BoT(D) & BoT(V) & BoT(SE) & BoT(Sh)  \\	
  					\hline
  					\hline
  					{Before Attack} & 75.90\% &	73.47\%&	66.39\%&	66.93\%&	66.92\%&	85.32\%&81.49\%&	76.74\%&	74.16\% &	 76.02\%\\ 
  					
  					\multirow{2}{*}{{After Attack}} & \multirow{2}{*}{1.15\% } &	\multirow{2}{*}{0.83\%}&	\multirow{2}{*}{0.71\%}&	\multirow{2}{*}{1.83\% }&	\multirow{2}{*}{0.71\% }&	\multirow{2}{*}{0.52\%}&	\multirow{2}{*}{0.52\%}&	\multirow{2}{*}{0.45\%}&	\multirow{2}{*}{0.78\%} &	 \multirow{2}{*}{0.19\%}\\ 
  					(mDR) &\textit{(98.49\%)}&\textit{(98.87\%)}&\textit{(98.93\%)}&\textit{(97.27\%)}&\textit{(98.94\%)}&\textit{(99.39\%)}&\textit{(99.36\%)}&\textit{(99.42\%)}&\textit{(98.95\%)}&\textit{(99.75\%)}\\
  					\hline
  				\end{tabular}
  			}
  			\label{bot}
  		\end{table*}
  	
  	\begin{table*}[!th]
  			\centering\scriptsize
  			\caption{\textbf{Accuracy (mAP) of extra state of the arts before and after attack on DukeMTMC-reID.} UAP is trained on BoT~\cite{bag-of-tricks} to attack these extra state of the arts. The corresponding \textit{mean mAP Drop Rate (mDR)} is also presented in parentheses.}\smallskip
  			\resizebox{1.0\textwidth}{!}{
  				\renewcommand{\arraystretch}{1.0}
  				\renewcommand{\tabcolsep}{0.0pt}
  				\begin{tabular}[c]{|C{2.6cm} || C{2.3cm} | C{2.2cm} | C{2.2cm} | C{2.2cm} | C{2.2cm} | C{2.2cm} |}
  					\hline
  					\multicolumn{1}{|c||}{\multirow{2}{*}{\small mAP}} & 
  					\multicolumn{3}{c|}{\textbf{Parted-based}}& \multicolumn{2}{c|}{\textbf{Attention-based}}& \textbf{Multi-clue} \\
  					\cline{2-7}
  					
  					& PCB & MGN & AlignedReID++ & SA-reID & HACNN &  St-reID   \\	
  					\hline
  					\hline
  					\multirow{2}{*}{{Before Attack}} & \multirow{2}{*}{74.27\%} &	\multirow{2}{*}{78.52\%}&	\multirow{2}{*}{72.86\%}&	\multirow{2}{*}{70.23\%}&	\multirow{2}{*}{63.23\%}&	\multirow{2}{*}{83.72\%}\\ 
  					&&&&&&\\
  					
  					\multirow{2}{*}{{Cross-framework attack}} & \multirow{2}{*}{4.68\% \textit{(93.69\%)}} &	\multirow{2}{*}{9.37\% \textit{(88.07\%)}}&	\multirow{2}{*}{0.86\% \textit{(98.82\%)}}&	\multirow{2}{*}{10.38\% \textit{(85.22\%)}}&	\multirow{2}{*}{13.82\% \textit{(78.14\%)}}&	\multirow{2}{*}{10.51\% \textit{(87.45\%)}}\\ 
  					&&&&&&\\
  					\multirow{2}{*}{{Within-framework attack}} & \multirow{2}{*}{0.84\% \textit{(98.87\%)}} &	\multirow{2}{*}{0.82\% \textit{(98.96\%)}}&	\multirow{2}{*}{0.29\% \textit{(99.60\%)}}&	\multirow{2}{*}{6.46\% \textit{(90.80\%)}}&	\multirow{2}{*}{10.65\% \textit{(83.16\%)}}&	\multirow{2}{*}{1.17\% \textit{(98.60\%)}}\\ 
  					&&&&&&\\
  					\hline
  					
  				\end{tabular}
  			}
  			\label{cros}
  		\end{table*} 
  	
  		\begin{table*}[!htb]
  			\centering\scriptsize
  			\caption{\textbf{The accuracy (mAP) of video-based Re-ID models before and after within-models MUAP attack.} The corresponding \textit{mean mAP Drop Rate (mDR)} is also presented in parentheses.}\smallskip
  			\resizebox{1.0\textwidth}{!}{
  				\renewcommand{\arraystretch}{0.8}
  				\renewcommand{\tabcolsep}{0.0pt}
  				\begin{tabular}[c]{| C{2.6cm} || C{2.5cm} | C{2.5cm} | C{2.5cm} | C{2.5cm} | C{2.5cm} | }	
  					\hline
  					\multirow{2}{*}{\small mAP} & 
  					\multirow{2}{*}{ResNet50~\cite{resnet50}} & \multirow{2}{*}{DenseNet121~\cite{densenet}} & \multirow{2}{*}{VGG19~\cite{vgg}} & \multirow{2}{*}{SeResNet50~\cite{senet}} & \multirow{2}{*}{ShuffleNet~\cite{shufflenet}}\\
  					& & & & & \\
  					\hline
  					\hline
  					\multirow{2}{*}{{Before Attack}}  & \multirow{2}{*}{82.93\%}& \multirow{2}{*}{83.03\%}& \multirow{2}{*}{78.37\%}& \multirow{2}{*}{75.25\%}&	\multirow{2}{*}{79.22\%}  \\
  					&&&&&\\
  					\multirow{2}{*}{{After Attack}}&  \multirow{2}{*}{1.31\% \textit{(98.42\%)}}&	\multirow{2}{*}{1.11\% \textit{(98.66\%)}}&	\multirow{2}{*}{2.30\% \textit{(97.07\%)}}&	\multirow{2}{*}{7.41\% \textit{(90.15\%)}}& \multirow{2}{*}{0.73\% \textit{(99.08\%)}} \\
  					&&&&&\\
  					\hline
  				\end{tabular}
  			}
  			\label{video-performance}
  		\end{table*}
  		
  		\begin{table*}[!thb]
  			\centering\scriptsize
  			\caption{\textbf{Attack on video-based Re-ID models.} UAP attack performance evaluation on MARS.}\smallskip
  			\resizebox{1.0\textwidth}{!}{
  				\renewcommand{\arraystretch}{1.0}
  				\renewcommand{\tabcolsep}{0.0pt}
  				\begin{tabular}[c]{| C{2.0cm} | C{1.2cm} || C{2.4cm} | C{2.4cm} | C{2.4cm} | C{2.4cm} | C{2.4cm} | }	
  					\hline
  					\multicolumn{2}{|c||}{mean mAP Drop Rate} & 
  					\multirow{2}{*}{ResNet50} & \multirow{2}{*}{DenseNet121} & \multirow{2}{*}{VGG19} & \multirow{2}{*}{SeResNet50} & \multirow{2}{*}{ShuffleNet}\\
  					\multicolumn{2}{|c||}{(mDR)} & & & & &\\
  					\hline
  					\hline
  					\multirow{2}{*}{ResNet50} & $\mathcal{L}_{AP}$ & \cellcolor{Gray!30}97.24\%& 45.18\%& 27.90\%& 10.00\%&	35.30\%  \\
  					& $+\mathcal{L}_{MI}$ &  \cellcolor{Gray!30}\textbf{98.42\%}&	\textbf{86.50\%}&	\textbf{71.21\%}&	\textbf{47.20}\%& \textbf{60.93\%} \\
  					\hline
  					\multirow{2}{*}{DenseNet121} & $\mathcal{L}_{AP}$ & 46.81\%& \cellcolor{Gray!30}98.22\%& 40.19\%&17.07\%&	52.34\% \\
  					& $+\mathcal{L}_{MI}$ &  \textbf{63.83\%}&	\cellcolor{Gray!30}\textbf{98.66\%}&	\textbf{65.42\%}& \textbf{38.91}\%&	\textbf{59.91\%}\\
  					\hline
  					\multirow{2}{*}{VGG19} & $\mathcal{L}_{AP}$ & 39.60\%& 43.02\%& \cellcolor{Gray!30}95.33\%&	26.25\%& 48.26\% \\
  					& $+\mathcal{L}_{MI}$ &  \textbf{52.79\%}&	\textbf{55.55\%}&	\cellcolor{Gray!30}\textbf{97.07\%}&	\textbf{37.63\%} & 56.01\%\\
  					\hline
  					\multirow{2}{*}{SeResNet50} & $\mathcal{L}_{AP}$ & 37.15\%& 44.49\%& 55.36\%&	\cellcolor{Gray!30}85.12\%& 53.23\% \\
  					& $+\mathcal{L}_{MI}$ &  \textbf{57.20\%}&	\textbf{59.80\%}&	\textbf{79.71\%}&	\cellcolor{Gray!30}\textbf{90.15\%} & 63.72\%\\
  					\hline
  					\multirow{2}{*}{ShuffleNet} & $\mathcal{L}_{AP}$ & 3.84\%& 5.99\%& 7.45\%&3.90\%&	\cellcolor{Gray!30}97.80\%  \\
  					& $+\mathcal{L}_{MI}$ &  \textbf{58.30\%}&	\textbf{60.04\%}&	\textbf{80.28\%}& \textbf{61.01}\%&	\cellcolor{Gray!30}\textbf{99.08\%}\\
  					\hline
  				\end{tabular}
  			}
  			\label{video-attack}
  		\end{table*}	
  		
	{\flushleft\textbf{Cross CNN architecture and dataset attack.}}
	In this subsection, we evaluate attack performance across CNN model and datasets, which is the most challenging protocol. We use the UAPs trained on the MSMT17 dataset to attack the models trained on Market-1501 and DukeMTMC-reID. The results in this scenario are reported in Table~\ref{cross-dataset-l2} using $\{\gamma=\infty, \epsilon=10, \lambda=10\}$ in terms of mDR.  The corresponding $RDR$ results are reported in Table~\ref{cross-dataset-rank1}.The results on the diagonal of the table marked in gray represent the performances in the special case where the source and target models share the same CNN architectures but with different parameters. It can be seen that the attack can remarkably disrupt the ranking list when the same CNN architecture is employed. However, similar to the previous experiment, the attack performance falls considerably when testing on a different CNN model. An interesting phenomenon is found in ShuffleNet where the learned UAP transfer better to different dataset than different CNN model. The reason may reside in that ShuffleNet is a lightweight CNN with much fewer parameters, which is helpful to avoid overfitting. However, its special operations such as channel shuffle are not used in other CNNs thus it performs much worse in the cross-model case. On the other hand, the proposed method with MI regularization increases the attack performance significantly in most cases.



	\subsection{{Attack on state-of-the-art image-based Re-ID models.}}
    \label{attacksota}
    We evaluate the performance of other person Re-ID models under UAP attack, including part-based models, attention-based models, and models which integrate spatial and temporal information. The performance of these models before and after the attack is reported in this part. Experimental results show that the proposed attack method significantly degenerates the performances of these models. 
  	{\flushleft\textbf{Detailed accuracy of BoT models before and after attack.}}
  	The attack transferability of MUAP cross CNN models or CNN architectures has been sufficiently explored in the paper using BoT~\cite{bag-of-tricks} based Re-ID framework trained with different backbones. Besides the mean mAP Drop Rate (mDR) reported in Table~\ref{single-dataset-l2} and Table~\ref{cross-dataset-l2} in the paper, we also present the detailed accuracy of these BoT~\cite{bag-of-tricks} models before and after attack in Table~\ref{bot} in terms of mAP.

  	\renewcommand{\tabcolsep}{1.0pt}
	\begin{figure*}[!th]
		\centering
		\small
		\begin{tabular}{cccccccccccccc}
			& & &\scriptsize{$\mathcal{L}_{AP}$} & \scriptsize{+$\mathcal{L}_{MI}$} & \scriptsize{$\mathcal{L}_{AP}$} & \scriptsize{+$\mathcal{L}_{MI}$} & \scriptsize{$\mathcal{L}_{AP}$} & \scriptsize{+$\mathcal{L}_{MI}$} & \scriptsize{$\mathcal{L}_{AP}$} & \scriptsize{+$\mathcal{L}_{MI}$} & \scriptsize{$\mathcal{L}_{AP}$} & \scriptsize{+$\mathcal{L}_{MI}$} \\
			{\rotatebox{90}{\small{DukeMTMC-reID}}} \hspace{1pt}&
			\includegraphics[width=0.078\linewidth]{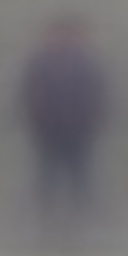} &
			\includegraphics[width=0.078\linewidth]{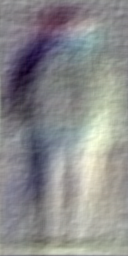} &
			\includegraphics[width=0.078\linewidth]{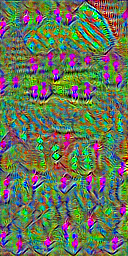} &
			\includegraphics[width=0.078\linewidth]{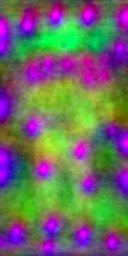} &
			\includegraphics[width=0.078\linewidth]{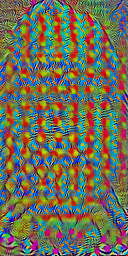} & 
			\includegraphics[width=0.078\linewidth]{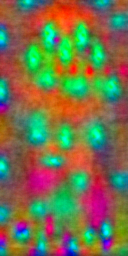} & 
			\includegraphics[width=0.078\linewidth]{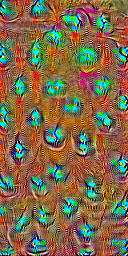}& 
			\includegraphics[width=0.078\linewidth]{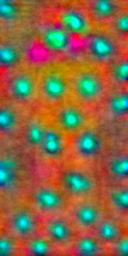} & 
			\includegraphics[width=0.078\linewidth]{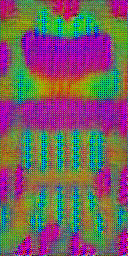} & 
			\includegraphics[width=0.078\linewidth]{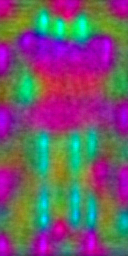} & 
			\includegraphics[width=0.078\linewidth]{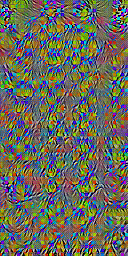} &
			\includegraphics[width=0.078\linewidth]{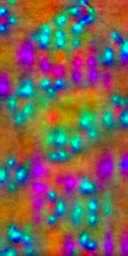} \\
			{\rotatebox{90}{~~~~~~\small{Market-1501}}} \hspace{1pt}&
			\includegraphics[width=0.078\linewidth]{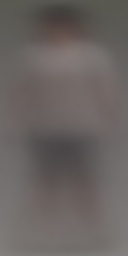} &
			\includegraphics[width=0.078\linewidth]{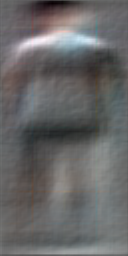} &
			\includegraphics[width=0.078\linewidth]{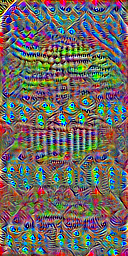} &
			\includegraphics[width=0.078\linewidth]{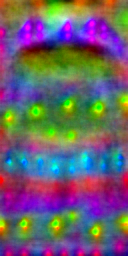} &
			\includegraphics[width=0.078\linewidth]{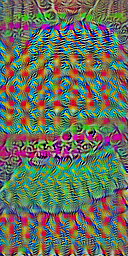} &
			\includegraphics[width=0.078\linewidth]{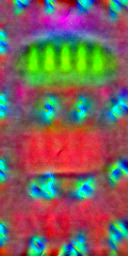} &  
			\includegraphics[width=0.078\linewidth]{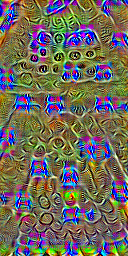} & 
			\includegraphics[width=0.078\linewidth]{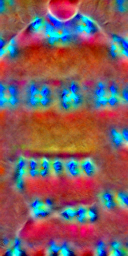}& 
			\includegraphics[width=0.078\linewidth]{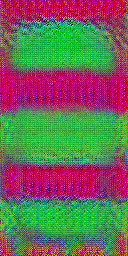} & 
			\includegraphics[width=0.078\linewidth]{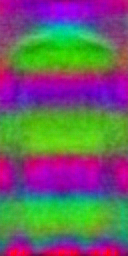} & 
			\includegraphics[width=0.078\linewidth]{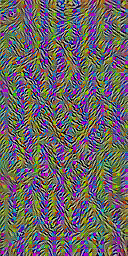} &
			\includegraphics[width=0.078\linewidth]{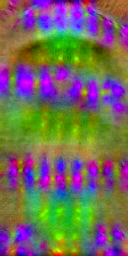} \\
			& \multicolumn{2}{c}{(a) Avg. img. \& grad.} & \multicolumn{2}{c}{(b) VGG-16} & \multicolumn{2}{c}{(c) ResNet-50} & \multicolumn{2}{c}{(d) DenseNet-121} & \multicolumn{2}{c}{(e) SENet-154} & \multicolumn{2}{c}{(f) ShuffleNet} \\
		\end{tabular}
		\caption{\textbf{UAPs trained with and without $\mathcal{L}_{MI}$ respectively on different models and datasets. \textit{It can be observed that there is a human body-like silhouette in the center of UAPs learned with $\mathcal{L}_{MI}$.} (a) Average images and gradients (normalized) of the training set on DukeMTMC-reID and Market-1501, respectively. (b)-(f) UAPs trained using different CNN architectures without and with $\mathcal{L}_{MI}$ regularizer. The pixel values are amplified ($10\times$) for visualization.}}
		\label{fig:moreuap_vis}
	\end{figure*}
  	{\flushleft\textbf{Cross Re-ID framework attack.}}
  	To further evaluate the attack performance on different Re-ID methods, we perform attack on extra state-of-the-art Re-ID frameworks, including part-based methods~\cite{mgn,alignedreid,pcb}, attention-based methods~\cite{SA,hacnn}, and multi-clue methods~\cite{st-ReID}. Part-based methods, such as MGN~\cite{mgn}, AlignedReID++~\cite{alignedreid} and PCB~\cite{pcb}, separate feature maps to several branches to learn discriminative local features. Attention-based methods, such as SA-reID~\cite{SA} and HACNN~\cite{hacnn}, utilize attention mechanism for better Re-ID performance. Multi-clue methods such as st-ReID~\cite{st-ReID} integrate spatial and temporal information. We perform cross-framework attack, i.e., we use UAP trained on BoT~\cite{bag-of-tricks} to attack these extra state of the arts. UAP is obtained in the same way as introduced in the paper. The accuracy of these methods before and after the cross-framework attack is reported in term of mAP in Table~\ref{cros}.
	 
	 As indicated in Table~\ref{cros}, models with all these frameworks can not defend the proposed MUAP attack. The accuracy of these state of the arts drops significantly. For example, the accuracy of AlignedReID++~\cite{alignedreid} on DukeMTMC-reID drops from 72.86\% to 0.86\% in term of mAP.  We further perform within-framework attack, The accuracy of these methods under within-framework attack is reported in Table~\ref{cros} (Row 3). The performance of cross-framework attack is very close to that of within-framework attack (Table~\ref{cros}, Row 2). Specifically, the difference of accuracy under cross-framework attack and within-framework attack is less than 10\% in terms of mAP, which indicates the high transferability of the MUAP attack across Re-ID methods.

  \subsection{{Attack on video-based person Re-ID models.}}
	\label{attackvideo}
    To sufficiently explore the robustness of Re-ID models, we further conduct experiments to attack video-based Re-ID models~\cite{video-reid}. As there are few works in video-based Re-ID with open source code, we extend BoT~\cite{bag-of-tricks} to video-based Re-ID using average pooling as the temporal modeling method~\cite{video-reid}. We train models with different backbones, including ResNet50~\cite{resnet50}, DenseNet121~\cite{densenet}, VGG19~\cite{vgg}, SeResNet50~\cite{senet}, and ShuffleNet~\cite{shufflenet}. We achieve performances comparable to state-of-the-art video-based Re-ID methods~\cite{video-occlusion,video-aanet} using ResNet50 and DenseNet121. The performances before and after attack is reported in Table~\ref{video-performance}. In video-based Re-ID attack, we first train a fixed UAP image, and then perform attack by adding the UAP to every frame of the Re-ID tracklets. The attack performances within and across models are reported in Table~\ref{video-attack}. As can be seen, our proposed method can significantly decrease the video-based Re-ID accuracy. Similar to the results in image-based Re-ID attack, the AP objective function (Eq.~\ref{eq:ap}) in the paper) is effective in disrupting the ranking list, and the MI regularization further improves the transferability of UAP across models.

	\subsection{Visualization}
	We visualize UAPs trained on different networks and different datasets without and with model-insensitive regularization in Fig.~\ref{fig:moreuap_vis}. 
	The perturbations trained with $\mathcal{L}_{MI}$ are apparently more smooth with less fine-grained noises than those trained without $\mathcal{L}_{MI}$, and share some similar patterns such as dots among different models. We also find that there always is a human body-like silhouette in the center of each perturbation trained with $\mathcal{L}_{MI}$ since minimizing $\mathcal{L}_{MI}$ has the effect on pushing perturbations into the common gradients of images. So that training UAPs with $\mathcal{L}_{MI}$ can learn the inherent structures of pedestrian images. Perturbations in the first and second row are trained on DukeMTMC-reID and Market-1501, respectively. As can be seen, the differences between perturbations trained on different CNN architectures are larger than those between perturbations trained on different datasets. This is consistent with our quantitative results in Fig.~4~(b) and (c) in the paper that cross CNN model attack is more challenging than cross-dataset attack in general.

	\subsection{Different forms of the model-insensitive regularization.} 
	\label{l1l2norm}
	In the paper, we propose a model-insensitive regularization which retains the natural image gradient distribution. To fully explore the potential of the regularization, we test different forms, including the $\ell_1$ and $\ell_2$ form. A general form of the MI regularization~(Eq.(10) in the paper) is formulated as follows:
	 	\begin{equation}
	 	\begin{aligned}
	 	\mathcal{L}_{MI} = &w_1 \sum_{i=1}^m{\left\|(\nabla q')_i\right\|_1} + w_2 \sum_{i=1}^m{{\left\|(\nabla q')_i\right\|_2}^2} \\ =& w_1 \sum_{i=1}^m{\left(\left\|(\partial_x q')_i\right\|_1 +  \left\|(\partial_y q')_i\right\|_1\right)} \\
	 	& + w_2 \sum_{i=1}^m{\left(\left\|(\partial_x q')_i\right\|_2^2 +  \left\|(\partial_y q')_i\right\|_2^2\right)},
	 	\end{aligned}
	 	\label{eq:generaltv}
	 	\end{equation}
	where $w_1$ and $w_2$ controls the effects of the $\ell_1$ and $\ell_2$ forms of regularization respectively. The experimental results are reported in Table~\ref{l1l2reg}. As can be seen, the $\ell_2$ form~($w_1=0, w_2 \neq 0$) is superior than the $\ell_1$ form~($w_1\neq0, w_2=0$) in cross-model attack. The reason may be that the $\ell_1$ regularization tends to cause sparse and peak perturbation signal which harms the transferability. Besides, the combination of $\ell_1$ and $\ell_2$ form~($w_1\neq0, w_2\neq0$) achieves similar transferability as $\ell_2$ form only in the best parameter setting. Therefore, we adopt the $\ell_2$ forms of regularization in the paper.
  		\begin{table}[!thb]
  			\centering\scriptsize
  			\caption{Parameter choosing for MI regularization. The elements in the table represents the cross-model attack performance~(mDR) with the corresponding prameter set. ($\epsilon=6$)}  \smallskip
  			\resizebox{0.45\textwidth}{!}{
  				\renewcommand{\arraystretch}{1.0}
  				\renewcommand{\tabcolsep}{0.0pt}
  				\begin{tabular}[c]{| C{1.0cm} || C{1.2cm} | C{1.2cm} | C{1.2cm} | C{1.2cm} | C{1.2cm} | }	
  					\hline
  					\diagbox{$w_2$}{$w_1$} & 0 & 0.01 & 0.1 & 1 & 10\\
  					\hline
  					\hline
  					0  & 13.54\%& 21.99\%& 27.35\%& 33.32\%&	11.24\%  \\
  					\hline
  					0.01 &  22.21\%& 22.22\%& 28.74\%& 32.19\% & 12.54\% \\
  					\hline
  					0.1 &  21.56\%& 24.81\%& 28.87\%&	33.37\%& 11.93\% \\
  					\hline
  					1 & 28.72\%& 30.47\%& 27.27\%&	33.03\%& 11.47\% \\
  					\hline
  					10 & 47.02\%& 46.93\%& 46.79\%&47.37\%&	5.91\%  \\
  					\hline
  					50 & 45.08\%& 44.48\%& 43.94\%& 39.74\%&	16.25\%  \\
  					\hline
  				\end{tabular}}
  			\label{l1l2reg}
  		\end{table} 	
  
	\section{Conclusion}
	This paper inspects the fragility of modern person Re-ID methods against dangerous universal adversarial attacks. We address the cross-model person Re-ID attack problem that existing UAP attack methods usually failed. We propose a MUAP person Re-ID attack method which is both model- and domain-insensitive. Extensive experimental results validate the proposed method. {We show that even in the challenging cross-model and cross-dataset attack scenario, where neither the Re-ID model nor the training dataset is available to the attacker, our proposed MUAP can decrease the Re-ID performance dramatically.} As a result, this study reveals that current Re-ID systems are susceptible to MUAP attacks and thus how to defend real-world person Re-ID systems accordingly should be seriously considered.
	
	\section*{Acknowledgment}
	This work is funded by National Key Research and Development Project of China under Grant No. 2019YFB1312000 and 2020AAA0105600, and by National Natural Science Foundation of China under Grant No. 62076195 and 62006183.
\ifCLASSOPTIONcaptionsoff
  \newpage
\fi



\bibliographystyle{IEEEtran}

\bibliography{bib/IEEEabrv.bib,bib/IEEEexample.bib}{}
\end{document}